\documentclass[10pt,twocolumn,letterpaper]{article}

 \usepackage{iccv}              
\usepackage[accsupp]{axessibility}
\definecolor{iccvblue}{rgb}{0.21,0.49,0.74}
\usepackage[pagebackref,breaklinks,colorlinks,allcolors=iccvblue]{hyperref}
\usepackage{amssymb}
\usepackage{multirow}
\usepackage{pifont}
\usepackage{bm}
\def\confName{ICCV}
\def\confYear{2025}

\title{GUIOdyssey: A Comprehensive Dataset for Cross-App GUI Navigation on Mobile Devices}

\author{Quanfeng Lu$^{4,1}$,\quad  Wenqi Shao$^{1,\dagger}$,\quad Zitao Liu$^{5}$,\quad Lingxiao Du$^{3,1}$,\quad Fanqing Meng$^{4,1}$, \\
Boxuan Li$^{3}$,\quad Botong Chen$^{5}$, \quad Siyuan Huang$^{4,1}$,\quad Kaipeng Zhang$^{1}$,\quad Ping Luo$^{2,\dagger}$\\
$^{1}$Shanghai AI Laboratory \quad $^{2}$The University of Hong Kong
\quad $^{3}$Nanjing University\\ $^{4}$Shanghai Jiao Tong University \quad  $^{5}$Harbin Institute of Technology, Shenzhen\\
{\tt\small shaowenqi@pjlab.org.cn, pluo@cs.hku.hk} \\
\small\url{https://github.com/OpenGVLab/GUI-Odyssey}
}

\begin{document}

\maketitle
\begingroup
  \renewcommand\thefootnote{$\dagger$}%
  \footnotetext{Corresponding author.}%
\endgroup
\begin{abstract}

Autonomous Graphical User Interface (GUI) navigation agents can enhance user experience in communication, entertainment, and productivity by streamlining workflows and reducing manual intervention. However, prior GUI agents often trained with datasets comprising tasks that can be completed within a single app, leading to poor performance in cross-app navigation. To address this problem, we present GUIOdyssey, a comprehensive dataset for cross-app mobile GUI navigation. GUIOdyssey comprises 8,334 episodes with an average of 15.3 steps per episode, covering 6 mobile devices, 212 distinct apps, and 1,357 app combinations. Each step is enriched with detailed semantic reasoning annotations, which aid the model in building cognitive processes and enhancing its reasoning abilities for complex cross-app tasks. Building on GUIOdyssey, we develop OdysseyAgent, an exploratory multimodal agent for long-step cross-app navigation equipped with a history resampler module that efficiently attends to historical screenshot tokens, balancing performance and inference speed. Extensive experiments conducted in both in-domain and out-of-domain scenarios validate the effectiveness of our approach. Moreover, we demonstrate that historial information involving actions, screenshots and context in our dataset can significantly enhances OdysseyAgent's performance on complex cross-app tasks.

\end{abstract}    
\section{Introduction}
\label{sec:intro}

Smartphones have become indispensable tools in our daily lives~\cite{barkhuus2011empowerment}. With a growing number of mobile applications, users frequently navigate across multiple apps to complete tasks, such as sharing content between social media platforms or coordinating schedules between messaging apps and calendars. Introducing a smart assistant to streamline these workflows and reduce manual intervention would be highly beneficial, particularly for individuals with physical disabilities~\cite{nanavati2023physically}. Nowadays, the rapid advancement of large foundation model~\cite{achiam2023gpt, dubey2024llama3herdmodels, Qwen2VL, chen2023internvl, Claude2023} has enabled the development of intelligent agents \cite{roziere2023code, achiam2023gpt, wang2024using, zhu2023ghost}. These agents process environmental observations, maintain multi-turn context, and execute actions to achieve specific goals, making autonomous GUI navigation increasingly feasible and practical.

\begin{figure*}
    \centering
    \includegraphics[width=0.75\textwidth]{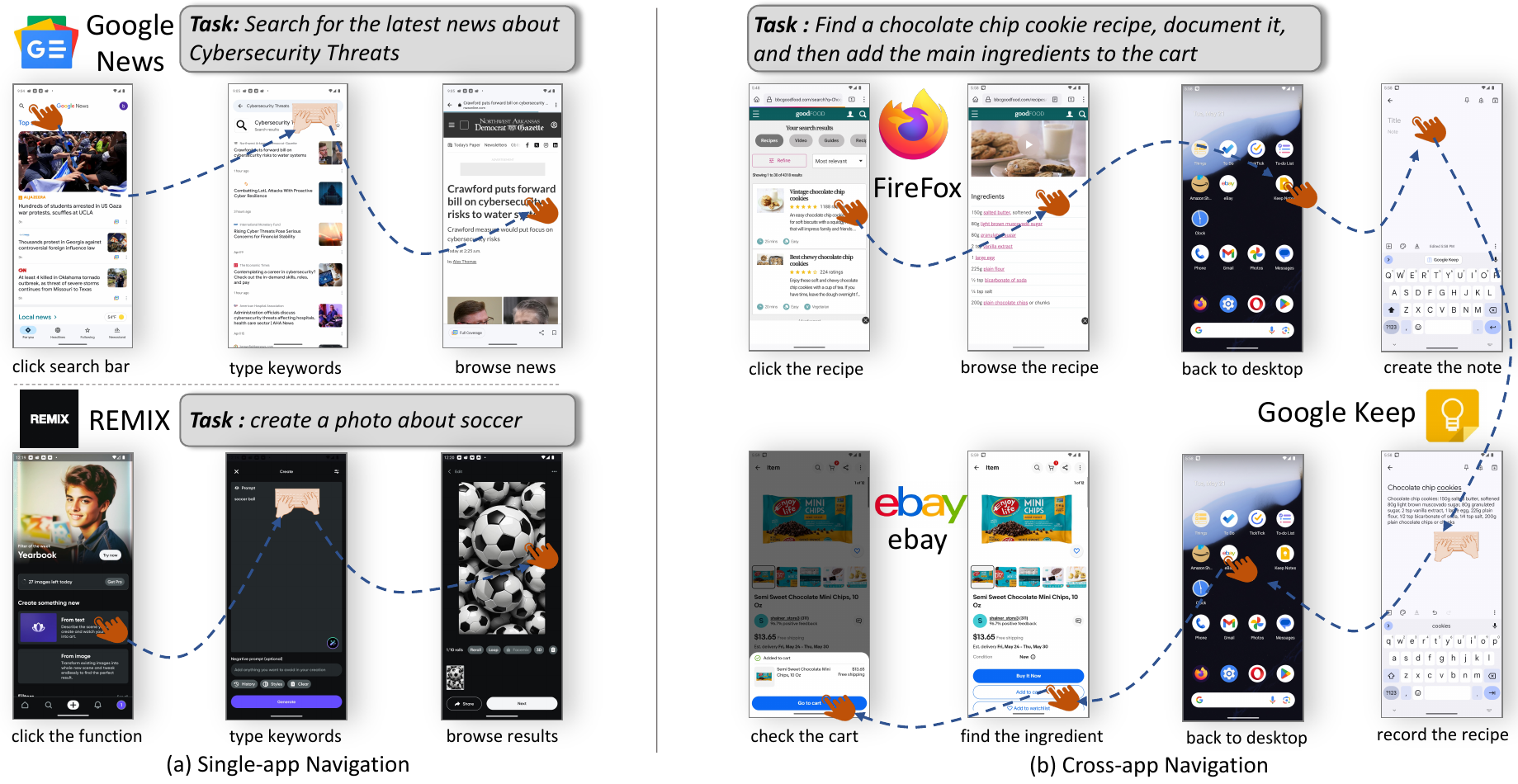}
    \caption{Illustration of single-app (a) and cross-app (b) GUI navigation. We see that cross-app navigation tasks demand the integration of multiple apps and the transfer of context and data between them, involving more complex workflows than single-app navigation.}
    \label{fig:single-cross-app}
\end{figure*}

While current foundation models are yet fully capable across various domains \cite{ying2024mmt}, they can still be effectively leveraged through GUI navigation datasets to build GUI agents that deliver more efficient and user-friendly mobile experiences. For instance, AITW \cite{rawles2023android} constructs a dataset encompassing various tasks to develop generalist agents for smartphones using large language models (LLMs) \cite{chowdhery2023palm}. Similarly, AndroidControl \cite{li2024effects} introduces a dataset focused on everyday tasks involving Android apps, providing both high-level and low-level instructions for GUI agents. 
However, these datasets primarily comprise operational actions, such as `click' and `scroll'. 
Furthermore, existing mobile GUI navigation datasets predominantly focus on tasks solvable within a single app, as depicted in \cref{fig:single-cross-app}(a). However, many real-world tasks require cross-app navigation, involving the transfer of context and data among multiple apps, as shown in \cref{fig:single-cross-app}(b). These complex workflows cannot be fully captured by single-app datasets, nor can they be decomposed without losing critical cross-app interactions.
While some studies have investigated cross-app tasks, their focus has been limited to evaluation purposes~\cite{rawles2024androidworld, xing2024understanding, li2024appagent, wang2024mobile2}. In particular, evaluations from studies~\cite{OSWorld, xing2024understanding} reveal that current performance on cross-app tasks remains significantly worse than on single-app tasks. Therefore, it is crucial to develop dedicated datasets to improve the cross-app navigation capabilities of GUI agents.

To address this issue and advance the development of general GUI agents, we introduce GUIOdyssey, the first cross-app GUI navigation dataset for mobile devices, featuring task instructions designed to reflect two levels of granularity. High-level instructions emulate natural human requests to capture real-world needs, whereas low-level instructions correspond to fine-grained tasks, providing precise and unambiguous guidance to eliminate potential misunderstandings. On one hand, we propose high-level \cite{rawles2023android, deng2024mind2web} cross-app navigation instructions by brainstorming with human participants and GPT-4 \cite{achiam2023gpt}, and create episode-specific user instructions to enrich task diversity. Independent annotators are then employed to annotate the entire navigation process comprehensively, including screenshots and corresponding actions, using an Android emulator\footnote{\url{https://developer.android.com/studio}}. On the other hand, after collecting the human-annotated data, we use GPT-4o \cite{gpt4o} to generate low-level instructions \cite{bai2021uibert, cheng2024seeclick} for each step, providing a more fine-grained guide to task completion and facilitating deeper exploration of the GUI agent's potential for cross-app tasks.

\begin{figure*}
    \centering
    \includegraphics[width=0.75\textwidth]{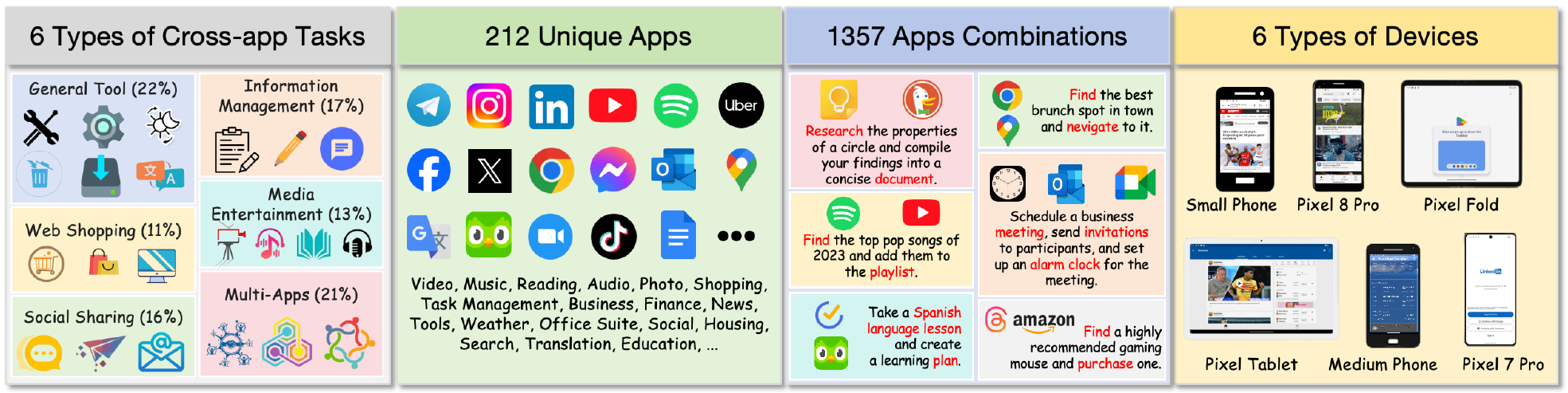}
    \caption{An overview of the proposed GUIOdessey. It encompasses $6$ types of cross-app navigation tasks spanning $212$ unique apps and $1,357$ combos of multiple apps from $6$ different devices.  }
    \label{fig:dataset-overview}
\end{figure*}

To simulate the human approach, we further enrich GUIOdyssey with semantic annotations by breaking down each step into three components: screen comprehension, historical context review, and decision-making reasoning. GPT-4o \cite{gpt4o} is employed to generate semantic annotations for these components. Subsequently, GUIOdyssey undergoes a quality check to ensure screenshot integrity, action accuracy, and alignment between GPT-4-generated instructions and the originals.
After construction through this rigorous pipeline, designed to enhance task diversity and annotation quality, GUIOdyssey comprises $8,334$ episodes with an average of $15.3$ steps, meticulously curated from $6$ different mobile devices such as Pixel Pro and Tablet. It features $6$ types of cross-app navigation tasks, ranging from general system tool use to media entertainment, involving navigation across $212$ different apps and $1,357$ app combinations in fields such as video, music, and reading, as depicted in \cref{fig:dataset-overview}. \cref{tab:dataset-compare} presents a comparison between GUIOdyssey and previous datasets.

Leveraging GUIOdyssey, we develop an exploratory cross-app multimodal agent named OdysseyAgent. Cross-app navigation tasks inherently involve long step sequences, requiring to retain numerous screenshots and actions for informed decision-making. However, processing large numbers of screenshot tokens can significantly slow inference, which is a critical concern for GUI agents frequently interacting with users. To balance performance with speed, OdysseyAgent incorporates a history resampler module that selectively attends to historical screenshot tokens while maintaining high inference throughput, thereby effectively and efficiently tackling complex cross-app tasks.
We thoroughly validate our approach on GUIOdyssey in both in-domain and out-of-domain scenarios. OdysseyAgent achieves highest accuracy among existing methods, including Claude3.5-Sonnet and GPT-4o. Moreover, incorporating semantic annotations leads to further performance gains. 
We also conduct an in-depth analysis demonstrating that enriching historical information with actions, screenshots, and contextual information significantly improves OdysseyAgent's performance, highlighting the importance of comprehensively modeling historical information for complex cross-app navigation tasks.

The contributions of this work are three-fold. 
1) We introduce GUIOdyssey, a comprehensive dataset for cross-app mobile GUI navigation, comprising $8,334$ episodes with an average length of $15.3$ steps. It covers a wide range of apps, tasks, and devices, with each step annotated by rich semantic reasoning to facilitate cognitive processes and enhance reasoning capabilities, thereby boosting performance on complex cross-app tasks. 2) We propose OdysseyAgent, an exploratory multimodal agent equipped with a history resampler module that balances performance and inference speed for cross-app navigation.
3) Through extensive experiments with OdysseyAgent, we demonstrate that comprehensively leveraging historical information substantially enhances performance on cross-app navigation tasks, highlighting the importance of historical information modeling.

\section{Related Work}

\

\textbf{GUI Navigation Agent.} 
Large foundation models \cite{achiam2023gpt, touvron2023llama, xu2023wizardlm, Qwen2VL, chen2023internvl, liu2024visual, Claude2023} have recently demonstrated the capacity to utilize extensive world knowledge to solve complex autonomous tasks \cite{yang2023mm, mu2024embodiedgpt, yao2022react, schick2024toolformer, shen2024hugginggpt}. These advancements have paved the way for the development of GUI agents capable of autonomous device control.
For instance, works such as \cite{gur2023real, deng2024mind2web, deng2024mind2web, zheng2024seeact} focus on autonomous agents in the Web domain, while studies like \cite{zhang2023appagent, li2024appagent, wang2024mobile, wang2024mobile2, yan2023gpt} leverage powerful language models, such as GPT-4V \cite{achiam2023gpt}, to address GUI navigation tasks on mobile devices. Additionally, other research \cite{wu2024copilot, zhang2024ufo} explores the potential applications of OS-specific agents.
This line of research often incorporates supplementary inputs, such as accessibility trees (A11y trees), to provide details like UI element coordinates or utilizes the Set-Of-Marks \cite{yang2023set} strategy to outline bounding boxes of UI elements, supported by GUI-specific grounding models \cite{lu2024omniparserpurevisionbased, gou2024uground}.
An alternative approach, as exemplified by \cite{shaw2023pixels, furuta2023multimodal, zhang2024android, hong2023cogagent, cheng2024seeclick, chai2024amex, you2025ferret, li2024ferret, wu2024atlas}, employs a coordinate-based method combined with visual models to develop GUI navigation agents. This approach directly provides positional information for executing actions, without relying on additional information.
While coordinate-based navigation can be fragile and may underperform in certain scenarios, it represents the ultimate solution for GUI navigation in the long run \cite{OSWorld}. In cases where structured A11y trees are unavailable or impractical \cite{OSWorld, cheng2024seeclick, shaw2023pixels}, coordinate-based navigation offers a natural and straightforward solution that enhances task and device transferability \cite{chen2024guicourse}. GUIOdyssey specifically adopts coordinate-based methods, aiming to create versatile, general-purpose GUI agents.

\textbf{Benchmarks and Datasets for GUI Agents.} Numerous benchmarks and datasets have been proposed to advance research in GUI navigation. Interactive online environments \cite{rawles2024androidworld, OSWorld, zhou2023webarena, koh2024visualwebarena, yao2022webshop, he2024webvoyager, pmlr-v70-shi17a, xing2024understanding} evaluate agents' GUI navigation capabilities, while other datasets \cite{bai2021uibert, cheng2024seeclick, you2025ferret, baechler2024screenai, liu2024visualwebbench, chen2024gui} primarily enhance UI perception and comprehension. Recent GUI datasets \cite{deng2024mind2web, kapoor2024omniact, lu2024weblinx, li2020mapping, venkatesh2022ugif, burns2021mobile, sun2022meta, rawles2023android, zhang2024android, chen2024guicourse, chai2024amex, li2024effects} predominantly involve tasks confined to a single app.
However, real-world usage frequently requires navigation across multiple apps, significantly increasing complexity. Cross-app tasks typically require longer action sequences (see \cref{tab:dataset-compare}), leading to higher error propagation risks. Additionally, cross-app interactions necessitate managing diverse working memory since key UI elements and contextual information span multiple apps. Furthermore, these tasks demand broader functional knowledge to integrate distinct interaction types like file sharing, email composition, and messaging. Additional examples of cross-app tasks are provided in Appendix \cref{sec:example}.
To address these challenges, we introduce GUIOdyssey, the first comprehensive cross-app GUI navigation dataset. A detailed comparison between GUIOdyssey and prior datasets is presented in \cref{tab:dataset-compare}.

\begin{table*}[t!]
\centering
\caption{GUI navigation dataset comparison. GUIOdyssey is a comprehensive cross-app GUI navigation dataset with over $8$k+ episodes, featuring an average of $15.3$ steps, which is the longest among mobile GUI datasets, and includes a diverse range of devices such as tablets.}
\label{tab:dataset-compare}
\scalebox{0.66}{
\begin{tabular}{l|ccccccc ccc}
\toprule

\textbf{Dataset} & \textbf{\# Episodes} & \textbf{\# Unique $\bm{I_{user}}$} & \textbf{\# Avg. Steps} & \textbf{Cross-app?} & \textbf{Platform} & \textbf{\# Domains}  & \textbf{Instruction Level} & \textbf{Semantic Annotation} \\

\midrule
Mind2Web ~\cite{deng2024mind2web} & 2,350 & 2,350  & 7.3  & \ding{55}  & Web   & 137 sites & high & \ding{55}  \\
WebLINX ~\cite{lu2024weblinx} & 2,337 & 2,337  & 43.0  & \ding{55}  & Web   & 155 sites & high & \ding{55}  \\

PixelHelp \cite{li2020mapping}  & 187 & 187  & 4.2 & \ding{55} & Phone & 4 apps & high \& low & \ding{55} \\
MoTIF \cite{burns2021mobile}  & 4,707 & 276  & 4.5 &\ding{55} & Phone & 125 apps  & high \& low & \ding{55}\\
UGIF \cite{venkatesh2022ugif} & 523 & 480  & 5.3 &\ding{55} & Phone & 12 apps & high \& low & \ding{51} \\
Meta-GUI & 4,684 & 1,125  & 5.3 &\ding{55} & Phone & 11 apps & high  & \ding{55} \\
AITW ~\cite{zhang2024android}    & 715,142 & 30,378  & 6.5 & \ding{55}  & Phone & 159 apps, 198+ sites & high & \ding{55}  \\
AITZ ~\cite{zhang2024android}    & 2,504 & 2,504 & 7.5  & \ding{55}  & Phone & 70+ apps & high & \ding{51}  \\
AndroidControl ~\cite{li2024effects}   & 15,283 & 14,548 & 5.5 & \ding{55}  & Phone & 833 apps & high \& low  & \ding{55}  \\
AMEX ~\cite{chai2024amex}   & 2,946 & 2,946 & 12.8 & \ding{55}  & Phone & 110 apps & high  & \ding{51}  \\
\midrule
\textbf{GUIOdyssey} & 8,334 & 8,334  & 15.3 & \ding{51} & Phone \& Tablet & 212 apps, 1,357 app combos & high \& low & \ding{51} \\
\bottomrule
\end{tabular}
}
\end{table*}

\section{GUIOdyssey Dataset}
  This section introduces the proposed cross-app navigation dataset. We present the metadata definition in \cref{sec:metadata-def}, details in data collection in \cref{sec:data-collection}, and dataset statistics in \cref{sec:data-stats}, respectively. The dataset overview is shown in \cref{fig:dataset-overview} and the collection process is presented in \cref{fig:data-pipeline}.

\subsection{Metadata Definition}\label{sec:metadata-def}
\begin{figure*}
    \centering
    \includegraphics[width=0.85\textwidth]{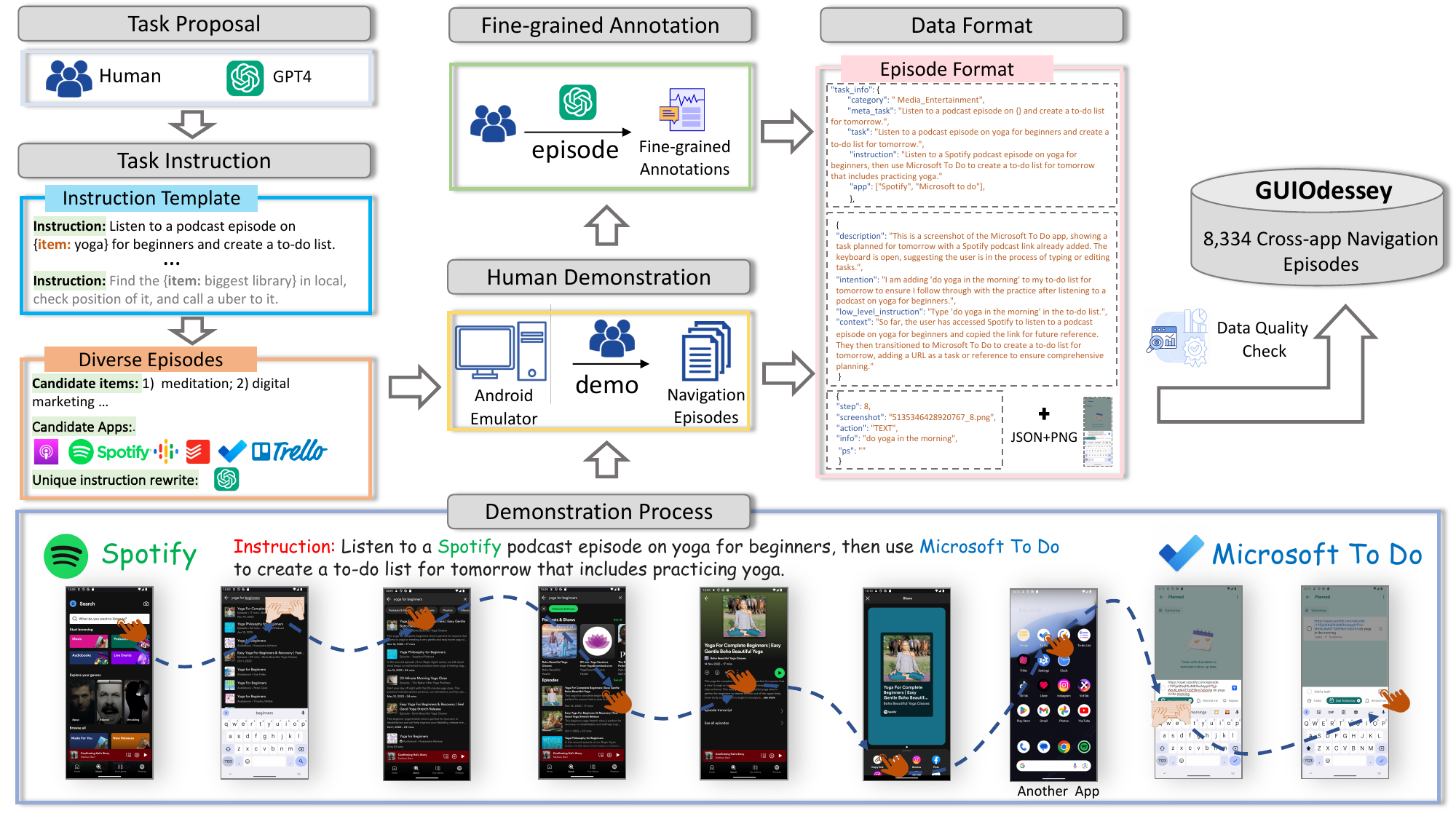}
    \caption{Data collection pipeline of GUIOdyssey for cross-app GUI navigation. GUIOdyssey comprises six categories of navigation tasks. For each category, we construct instruction templates with items and apps selected from a predefined pool, resulting in a vast array of unique instructions for annotating GUI episodes. Human demonstrations on an Android emulator capture the annotation of each episode in a comprehensive format. After rigorous quality checks, GUIOdyssey includes $8,334$ validated cross-app GUI navigation episodes.}
    \label{fig:data-pipeline}
\end{figure*}

\
\textbf{GUI Episode.} A GUI episode is a recorded sequence of interactions capturing the action steps to complete the navigation task from the user's high-level instruction. Formally, given the user's high-level instruction $I_{user}$ and the screenshot $X^t$ at the time step $t$, the GUI Agent $\mathcal{G}$ will take the action $A^t=\mathcal{G}(X^t, I_{user})$ to complete this instruction. When the task is completed, the episode is defined as the sequence including all screenshots and actions denoted as ${E}=\{(X^t, A^t)_{t=1}^T, I_{user}\}$ where $T$ indicates the total steps. An example of the episode is illustrated in \cref{fig:data-pipeline}. Note that the total step $T$ of cross-app navigation is much larger than that of single-app navigation as shown in \cref{fig:single-cross-app}.

\textbf{Action Set.} The action set of GUIOdyssey comprises $9$ kinds of actions: \texttt{CLICK, SCROLL, LONG PRESS, TYPE, COMPLETE, IMPOSSIBLE, HOME,  BACK}, and \texttt{RECENT}. The arguments and functionalities of these actions are summarized in \cref{tab:action-space} of Appendix \cref{sec:app-action-set}. 

\subsection{Data Collection}\label{sec:data-collection}
\
\textbf{Cross-app Task Proposal.} 
As depicted in \cref{fig:dataset-overview}, GUIOdyssey comprises six types of cross-app navigation tasks: 1) \textbf{General Tool}, which includes tasks that entail system-wide operations. 2) \textbf{Information Management}. It encompasses the activities of searching for and recording information for future utilization. 3) \textbf{Web Shopping}. Shopping encompass a variety of activities associated with online product purchases. 4) \textbf{Media Entertainment}, which revolves around engaging in activities related to video and music streaming applications. 5) \textbf{Social Sharing} encompasses activities where users share content across various social media platforms, and 6) \textbf{Multi-Apps}, which involve more complex operations across different domains. See Appendix~\cref{sec:desc-gui-odyssey} for details on these tasks.

\textbf{High-Level Task Instruction.} For all aforementioned cross-app tasks, we propose a flexible high-level instruction template to construct diverse GUI episodes. The instruction templates are generated by i) human participants and ii) prompting GPT-4 with task descriptions. Ultimately, we collect $91$ high-level instruction templates. The diversity of instructions is implemented in three ways. First, the item in each template can be replaced with various candidates. For instance, the item in the instruction ``Listen to a podcast episode on \{item: yoga\} for beginners and create a to-do list" can be substituted with ``meditation" or ``digital marketing" as shown in \cref{fig:data-pipeline}. 
Second, the apps used to complete the instruction can be selected from a predefined pool. For example, the podcast app can be Spotify or Google Podcast and the scheduling app can be Todoist or Microsoft To Do.  Finally, we employ GPT-4 to rewrite the instruction using candidate items and apps with different expressions.

\textbf{Human Demonstration.}
With diverse high-level instructions collected, we then engage independent annotators, experienced in using mobile devices and various apps, participate in the annotation of GUI episodes. As mentioned in \cref{sec:metadata-def}, we use an Android emulator to record GUI episodes on various mobile devices such as Pixel Pro, Tablet, and Fold as shown in \cref{fig:dataset-overview}. All annotators are required to complete the instructions step-by-step and avoid clicking on anything unrelated to the task while recording their interactions. To improve data quality, annotators are trained to annotate at least twenty episodes before starting annotation. During annotation, annotators are asked to save the screenshot before each action step. As shown in \cref{tab:action-space} of Appendix \cref{sec:app-action-set}, we use the actions \texttt{IMPOSSIBLE} and \texttt{COMPLETE} to denote the instructions that cannot be completed and those that have been completed, respectively. Specifically, when annotators select \texttt{IMPOSSIBLE}, they are required to record the reason why the task could not be completed. Upon completion of the navigation, our data annotation tools save the episode, including the user's instructions, screenshots, actions taken at each step, the apps used by the annotator, and any additional notes.
An example of the annotation process is illustrated in \cref{fig:data-pipeline}.

\textbf{Fine-grained Episode Annotation.}
After collecting human-demonstrated GUI episodes, we utilize the state-of-the-art model GPT-4o to generate fine-grained episode annotations, consisting of two main components. The first component is the \textbf{Low-Level Instruction}, which refers to a set of fine-grained instructions that serve as atomic decompositions of high-level instructions, providing detailed steps for executing the next action on the current page. The second component is \textbf{Semantic Annotation}, which includes: (1) \textbf{Screen Description}, offering a detailed depiction of the content displayed in the screenshot; (2) \textbf{Contextual Information}, summarizing the preceding steps that led to the current stage of the task; and (3) \textbf{Decision Rationale}, explaining the reasoning behind the next action based on both historical context and the current screen content.  Further details can be found in Appendix \cref{sec:semantic_detail}, while an example of the semantic annotation process is illustrated in \cref{fig:data-pipeline}.

\textbf{Data Quality Check.}
With all episodes collected, we perform a data quality check. The episode is thought to be accurate and complete if it satisfies the following three criteria: i) whether any screenshot in the episode is regularly saved; ii) whether the sequence of screenshots and actions can complete the instruction; iii) whether the instruction rewritten by GPT-4 is equivalent to the original one. After filtering low-quality data, we obtain our cross-app navigation dataset called GUIOdyssey.

\subsection{Dataset Statistics}\label{sec:data-stats}

\begin{figure*}[h]
    \centering
    \includegraphics[width=0.8\textwidth]{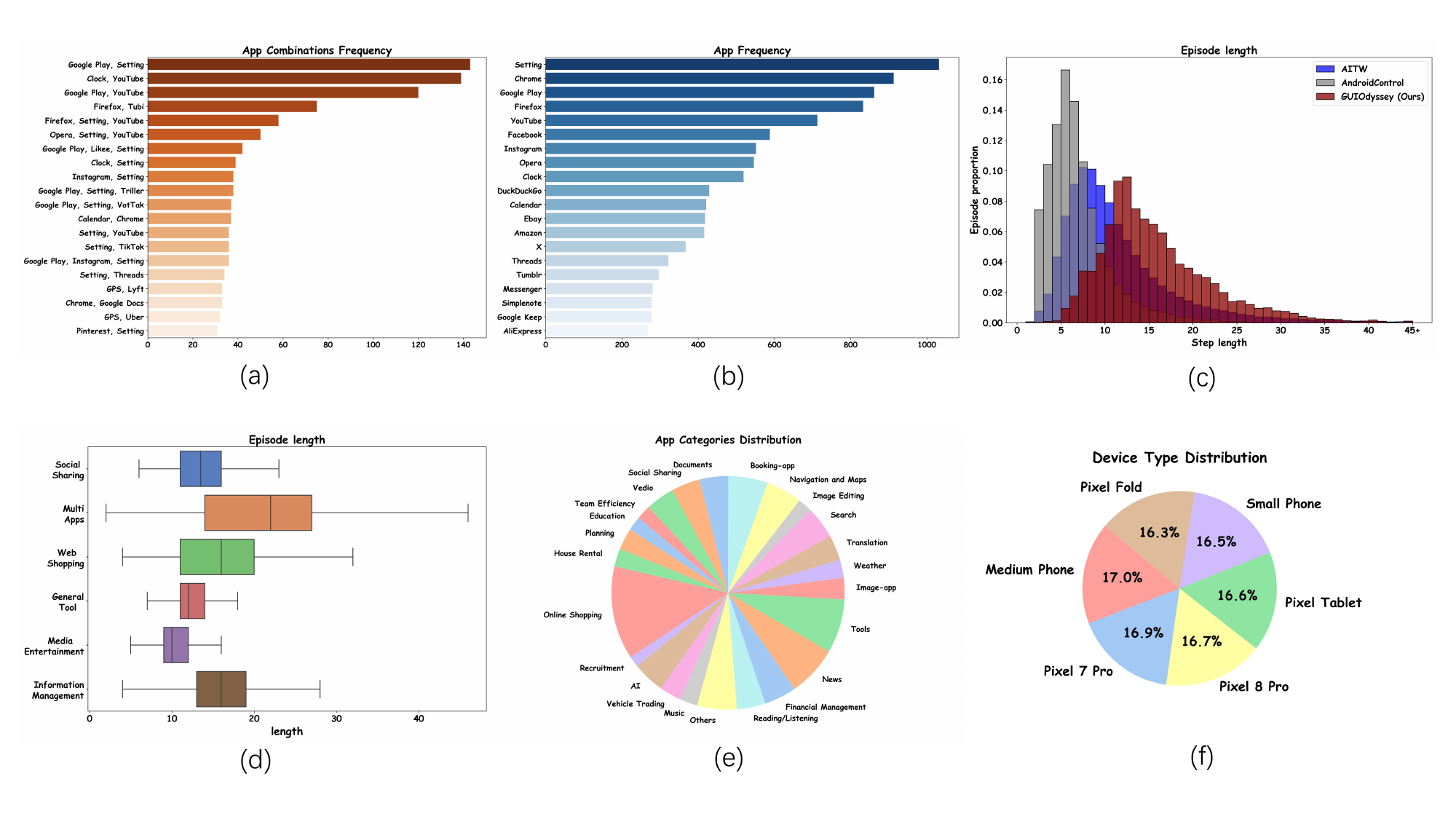}
    \caption{Statistics for GUIOdyssey, zoom in to view details. (a) App Combinations Frequency. (b) App Frequency. (c) Episode length of AITW, AndroidControl and GUIOdyssey. (d) Episode length distribution. (e) App categories distribution. (f) Device statistics.}
    \label{fig:stat}
\end{figure*}

GUIOdyssey targets cross-app navigation, a more practical scenario than single-app navigation in real-world settings. It comprises $8,334$ episodes with an average of $15.3$ steps per episode, making it the mobile GUI navigation dataset with the longest average episode length. Compared to existing datasets, GUIOdyssey encompasses a broader range of navigation tasks and more complex workflows, featuring six types of cross-app tasks that span $212$ apps across domains such as video, music, and reading. It also includes six types of electronic devices, including foldable phones and tablets, which were not covered in previous datasets. Visual statistics are presented in \cref{fig:stat}, where \cref{fig:stat} (c) highlights its significantly longer episode lengths compared to single-app datasets\cite{rawles2023android, li2024effects}. Other provide additional insights into app combination and usage frequency (\cref{fig:stat} a, b), episode length distribution across task types (\cref{fig:stat} d), the presence of $25$ app categories (\cref{fig:stat} e), and the diversity of device types (\cref{fig:stat} f).

\section{Method: OdysseyAgent}\label{sec:cross-app-agent}

Building upon GUIOdyssey, we introduce OdysseyAgent, an exploratory framework for cross-app navigation tasks powered by Large Vision-Language Models (LVLMs). 
A key challenge in cross-app tasks is balancing the need to process numerous historical screenshots and lengthy action sequences with the requirement for fast inference in frequent user interactions. To address these demands, we fine-tune Qwen-VL \cite{bai2023qwen} on GUIOdyssey, incorporating a history replay module to optimize both performance and efficiency.

\begin{figure}[h!]
    \centering
    \includegraphics[width=0.85\linewidth]{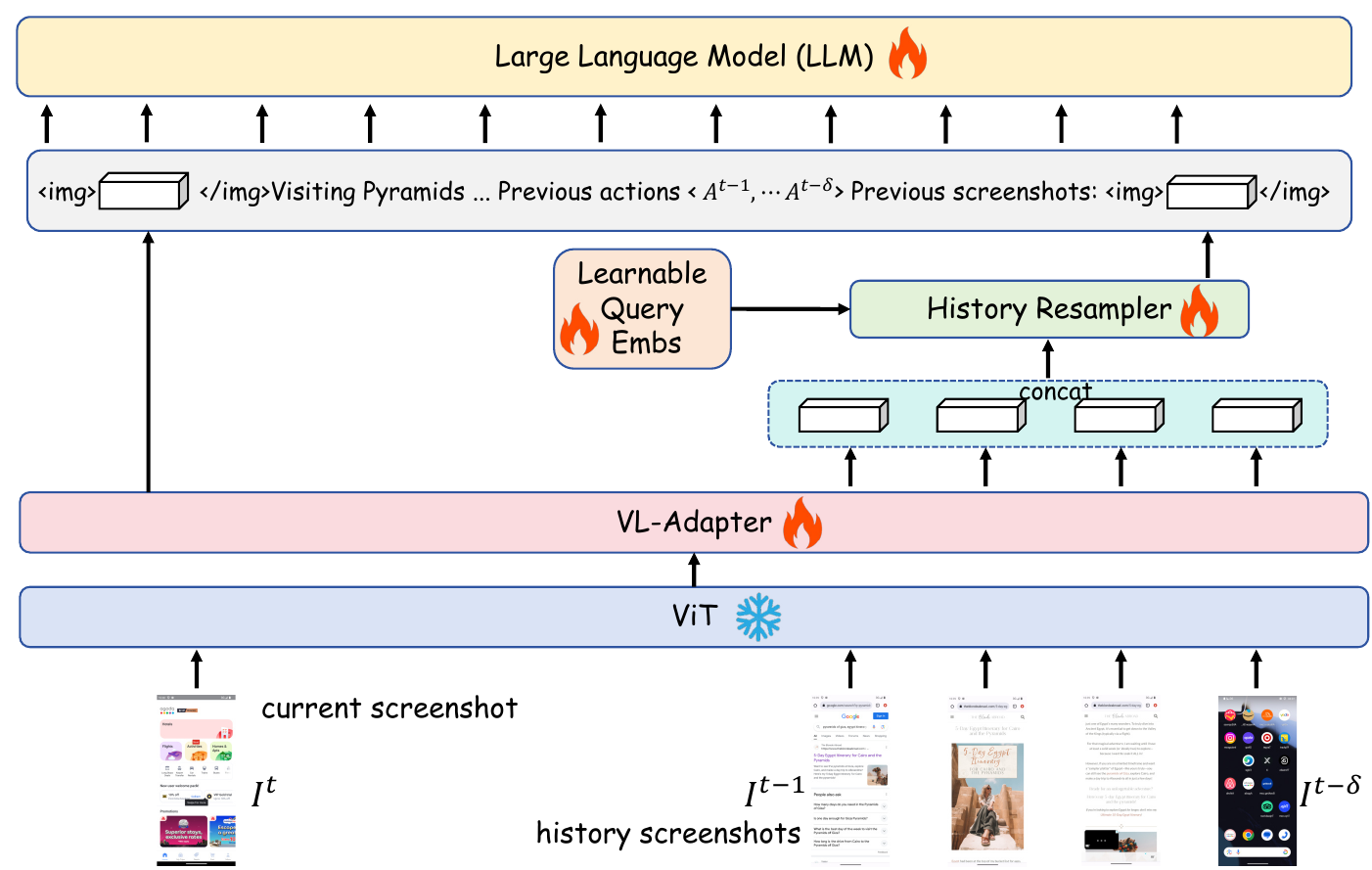}
    \caption{The architecture of OdysseyAgent. Beyond Qwen-VL's standard components, OdysseyAgent introduces a history resampler that enables efficient attention to historical screenshots.}
    \label{fig:odyssey-agent}
\end{figure}

As illustrated in \cref{fig:odyssey-agent}, OdysseyAgent inherits from Qwen-VL-Chat \cite{bai2023qwen} and comprises a vision encoder, a large language model (LLM), and a vision-language (VL) adapter. Crucially, we introduce a history resampler to compress historical screenshot tokens before they reach the LLM. This design alleviates the overhead of stacking all past screenshots while still leveraging essential contextual information.
In Appendix \cref{sec:strategy_vs}, we compare the history resampler with a straightforward multi-image concatenation approach, demonstrating that the history resampler achieves a more favorable balance between performance and inference efficiency.

Specifically, the history resampler is implemented as a single-layer cross-attention module, where learnable embeddings serve as the query and historical screenshot tokens function as both key and value. After resampling, the compressed historical screenshot tokens are concatenated with the current screen image token, user instruction, and previous actions. This fused representation is then fed into the LLM to predict the next action. Formally, the next-word prediction objective $\mathcal{L}$ is defined as: $\mathcal{L}=\sum_{i=1}^N P_\theta(A^t_i|X^{\{t,t-1,\cdots,t-\delta\}},I_{user},A^t_{<i})$, where $N$ is the number of tokens in action $A^t$, $\delta$ denotes the historical image window, and $\theta$ represents the trainable parameters in OdysseyAgent (namely the VL adapter, history resampler, and LLM as shown in \cref{fig:odyssey-agent}).

\section{Experiment}
The experimental setup is detailed in \cref{sec:exp-setup}. In \cref{sec:exp-results-dyssey}, we evaluate OdysseyAgent's performance under both in- and out-of-domain settings. \cref{sec:exp-ablation} further explores the role of historical information in cross-app tasks.

\subsection{Experimental Setup}\label{sec:exp-setup}

We leverage the comprehensiveness of GUIOdyssey to evaluate OdysseyAgent’s performance in both in- and out-of-domain scenarios. To this end, we divide GUIOdyssey into four distinct setups. The first is an in-domain split: (i) \textbf{Train-Random \& Test-Random}. The remaining three are out-of-domain splits: (ii) \textbf{Train-App \& Test-App}, (iii) \textbf{Train-Task \& Test-Task}, and (iv) \textbf{Train-Device \& Test-Device}. These setups are designed to assess the agent's generalizability across different app, task, and device scenarios. A detailed description of the four setups is provided in Appendix \cref{sec:split-desc}, while the training details are available in \cref{sec:training_datail}.

\begin{table*}[t!]
\centering
\caption{Results of different LVLMs on Test-Random split. The evaluation metric is the action matching score (AMS). `HL' and `LL' indicate that the task instruction is high-level and low-level, respectively. ${}^{\ast}$ indicates that agent's training also includes semantic annotations.}
\label{tab:main_in}
\scalebox{0.67}{
\begin{tabular}{l|cc| cc| cc| cc|cc|cc|cc }
\toprule
\multirow{2}*{\textbf{Model}}  & \multicolumn{2}{c|}{\textbf{Tool}} & \multicolumn{2}{c|}{\textbf{Information}} & \multicolumn{2}{c|}{\textbf{Shopping}} & \multicolumn{2}{c|}{\textbf{Media}} & \multicolumn{2}{c|}{\textbf{Social}} & \multicolumn{2}{c|}{\textbf{Multi-Apps}} & \multicolumn{2}{c}{\textbf{Overall}} \\
& HL & LL & HL & LL & HL & LL & HL & LL & HL & LL & HL & LL & HL & LL \\

\midrule
\multicolumn{15}{c}{\textit{zero-shot}} \\
GPT-4V & 14.93 & 40.86 & 14.69 & 38.56 & 12.17 & 36.04 & 10.80 & 48.40 & 16.79 & 43.21 & 11.54 & 40.61 & 13.49 & 41.28  \\
GPT-4o & 14.15 & 38.11 & 13.86 & \textbf{42.40} & 11.69 & 40.57 & 12.00 & 52.40 & \textbf{18.14} & \textbf{44.44} & 9.28 & 38.35 & 13.19 & 42.71  \\
Claude3.5-sonnet & \textbf{22.99} & 40.28 & 14.69 & 34.56 & 12.17 & 31.03 & 14.00 & 37.60 & 16.79 & 30.62 & 14.14 & 31.00 & 15.80 & 34.18 \\
InternVL2-Pro & 19.45 & \textbf{49.51} & 15.86 & 40.23 & \textbf{17.18} & \textbf{41.05} & 13.20 & \textbf{51.60} & 14.81 & 40.00 & \textbf{15.72} & \textbf{41.52} & \textbf{16.04} & \textbf{43.98} \\
CogAgent & 18.81 & 33.52 & 12.35 & 29.79 & 13.02 & 26.89 & 12.63 & 25.80 & 14.72 & 33.54 & 12.15 & 33.09 & 13.95 & 30.44 \\
SphAgent & 22.24 & 36.81 & \textbf{17.43} & 29.95 & 13.60 & 25.08 & \textbf{15.54} & 33.03 & 14.20 & 28.83 & 12.89 & 26.32 & 15.98 & 30.00 \\
\midrule
\multicolumn{15}{c}{\textit{zero-shot with OmniParser}} \\
GPT-4V & 24.37 & 56.03 & 22.44 & 51.10 & 17.16 & 46.75 & 18.65 & 62.18 & 32.28 & 59.18 & 24.21 & 54.58 & 23.18 & 54.97 \\
GPT-4o &  26.63 & 55.28 & 23.45 & 53.91 & 18.34 & 47.63 & 19.17 & 63.21 & 31.01 & 61.08 & 23.39 & 55.95 & 23.67 & 56.18  \\
Claude3.5-sonnet & \textbf{39.20} & \textbf{64.07} & \textbf{28.46} & \textbf{61.92} & \textbf{27.22} & \textbf{56.51} & \textbf{28.50} & \textbf{66.84} & \textbf{40.82} & \textbf{68.99} & \textbf{33.11} & \textbf{65.12} & \textbf{32.88} & \textbf{63.91} \\
InternVL2-Pro & 16.58 & 58.04 & 16.03 & 51.30 & 8.88 & 49.70 & 16.58 & 60.62 & 16.14 & 53.80 & 13.95 & 52.39 & 14.69 & 54.31  \\
\midrule
\multicolumn{15}{c}{\textit{fine-tuned}} \\
Qwen-VL & 85.55 & 90.79 & 68.04 & 83.36 & 62.28 & 80.67 & 77.56 & 88.15 & 80.29 & \textbf{88.36} & 74.27 & 86.56 & 74.67 & 86.32\\
${\text{Qwen-VL}}^{\ast}$ & 86.35 & 90.99 & \textbf{72.01} & 85.77 & 67.31 & 82.86 & 80.33 & 89.48 & 82.39 & 88.01 & 77.52 & \textbf{89.40} & 77.65 & 87.78 \\
\textbf{OdysseyAgent} & 86.01 & 91.21 & 69.83 & 83.37 & 65.19 & 82.63 & 77.10 & 88.55 & 81.47 & 87.66 & 75.13 & 87.84 & 75.79 & 86.88  \\

${\textbf{OdysseyAgent}}^{\ast}$ & \textbf{86.82} & \textbf{91.25}  & 71.79  & \textbf{86.58}  & \textbf{68.58}  &  \textbf{83.74} & \textbf{80.93}  &  \textbf{89.66} &  \textbf{82.88} & 88.27  & \textbf{78.47}  & 89.39  &  \textbf{78.24} &  \textbf{88.15} \\
\bottomrule
\end{tabular}
}
\end{table*}

\begin{table}[t!]
\centering
\caption{OdyssseyAgent's performance on out-of-domain tasks. `Semantic?' indicates whether the model's training includes semantic annotations. `HL' and `LL' indicate that the task instruction is high-level and low-level, respectively.}
\label{tab:main_ood}
\scalebox{0.64}{
\begin{tabular}{c|c|cc cc cc|cc}
\toprule
\multirow{2}*{\textbf{Semantic?}} & \textbf{Task}& \multicolumn{2}{c}{\textbf{Test-Task}} & \multicolumn{2}{c}{\textbf{Test-Device}}  & \multicolumn{2}{c|}{\textbf{Test-App}} & \multicolumn{2}{c}{\textbf{Overall}} \\ 
& \textbf{Level} &  AMS & SR &  AMS & SR & AMS & SR & AMS & SR  \\ 
\midrule
\multirow{2}*{\ding{55}} & HL & 54.36 & 0.09 & 61.20 & 1.88 & 63.03 & 7.70 & 59.53 & 3.22 \\
& LL & 78.97 & 2.20 & 79.66 & 8.47 & 84.24 & 20.70 & 80.96 & 10.46 \\
\midrule
\multirow{2}*{\ding{51}} & HL & 56.19 & 0.26 & 66.63 & 5.07 & 65.89 & 8.81 & 62.90 & 4.71 \\
& LL & 80.19 & 2.29 & 79.93 & 11.66 & 83.47 & 20.02 & 81.20 & 11.32 \\
\bottomrule
\end{tabular}
}
\end{table}

\begin{table*}[h!]
\centering
\caption{The impact of different historical components in GUIOdyssey across four splits. High-level instructions are used for both training and evaluation, and performance is measured by AMS and SR.}
\label{tab:historical_comparison}
\scalebox{0.80}{
\begin{tabular}{c|ccc| cc | cc | cc | cc|cc}
\toprule

 & \multicolumn{3}{c|}{\textbf{Historical Information}} & \multicolumn{2}{c|}{\textbf{Test-Random}} & \multicolumn{2}{c|}{\textbf{Test-Task}}  & \multicolumn{2}{c|}{\textbf{Test-Device}} & \multicolumn{2}{c|}{\textbf{Test-App}} & \multicolumn{2}{c}{\textbf{Overall}}  \\ 
 & action & screenshot & context & AMS & SR & AMS & SR & AMS & SR & AMS & SR & AMS & SR  \\
\midrule
(1) & \ding{55} & \ding{55} & \ding{55} & 66.13 & 1.65 & 47.62 & 0.00 & 54.15 & 0.72 & 54.49 & 3.59 & 55.60 & 1.49 \\
\midrule
(2) & \ding{51} & \ding{55} & \ding{55} & 74.67 & 9.70 & 55.00 & 0.00 & 62.03 & 2.03 & 62.06 & \textbf{8.98} & 63.44 & 5.18 \\
(3) & \ding{55} & \ding{51} & \ding{55} & 71.22 & 6.69 & 51.69 & 0.09 & 59.12 & 2.24 & 59.16 & 7.78 & 60.30 & 4.20 \\
(4) & \ding{55} & \ding{55} & \ding{51} & 75.25 & 9.50 & 57.66 & \textbf{0.62} & 62.35 & 2.24 & 63.82 & 7.87 & 64.77 & 5.06 \\
\midrule
(5) & \ding{51} & \ding{51} & \ding{55} & 75.79 & 9.38 & 54.36 & 0.09 & 61.20 & 1.88 & 63.03 & 7.70 & 63.60 & 4.76 \\
\midrule
(6) & \ding{51} & \ding{51} & \ding{51} & \textbf{77.06} & \textbf{11.61} & \textbf{58.83} & 0.18 & \textbf{65.85} & \textbf{5.00} & \textbf{65.63} & 8.47 & \textbf{66.84} & \textbf{6.32} \\
\bottomrule
\end{tabular}
}
\end{table*}

\textbf{Evaluation Metrics.} To ensure reproducibility and efficiency, we adopt an offline evaluation method to benchmark performance. We use the \textbf{Action Matching Score (AMS)} as our metric, inspired by the approaches presented in AITW \cite{rawles2023android} and AutoUI \cite{zhan2023you}. An action is considered correct if its action type matches the ground-truth type. Additionally, for \texttt{CLICK} and \texttt{LONG PRESS} actions, we consider them correct if they fall within $14\%$ of the screen distance from the reference gesture. Furthermore, we utilize SAM2 \cite{ravi2024sam2} to determine the coordinates of the target element, and if the predicted coordinates lie within the region segmented by SAM2, the action is also deemed correct. As for \texttt{SCROLL} actions, we compare whether the direction (\emph{i.e.}, up, down, left, or right) matches the gold gesture's direction. For \texttt{TYPE} actions, we evaluate the Average Normalized Levenshtein Similarity (ANLS) \cite{biten2019scene} between the predicted and gold gestures. If the ANLS is below a certain threshold (set to $0.5$ in our experiments), we consider it correct. We then calculate \textbf{Success Rate (SR)} for the whole episode. A task is considered successful only if all actions are correct. Success Rate (SR) is a rigorous metric. It would be harder to achieve higher SR in tasks with more action steps.

\subsection{Comprehensive evaluation on the GUIOdyssey}\label{sec:exp-results-dyssey}

We evaluate OdysseyAgent's performance in both in-domain and out-of-domain scenarios. For each step in the dataset, we construct prompts using high-level and low-level instructions separately for training and evaluation. High-level instructions reflect the model's capability to handle real-world GUI navigation tasks, while low-level instructions break down each step of the high-level tasks, assessing the model's ability to follow simpler commands. Naturally, high-level instructions are more challenging than low-level instructions.

\textbf{In-domain Performance.}
We compare OdysseyAgent against three types of methods on the Test-Random split of GUIOdyssey: (1) LVLMs zero-shot, including closed-source proprietary LVLMs (GPT-4V \cite{achiam2023gpt}, GPT-4o \cite{gpt4o}, Claude3.5-Sonnet \cite{Claude2023}, InternVL2-Pro \cite{chen2024far}) and open-source GUI-specific models (SphAgent \cite{chai2024amex}, CogAgent \cite{hong2023cogagent}); (2) closed-source LVLMs zero-shot with OmniParser \cite{lu2024omniparserpurevisionbased}; and (3) fine-tuned LVLMs Qwen-VL\cite{bai2023qwen}.
Due to budget constraints, for closed-source models, we sample $200$ episodes from the original test set to serve as their evaluation set.
Note that Qwen-VL is effectively OdysseyAgent without the history resampler, meaning it does not incorporate historical screenshots.
The result is shown in \cref{tab:main_in}. 
InternVL2-Pro achieves the best overall performance among all coordinated-based models. Despite being trained on other GUI navigation datasets, CogAgent and SphAgent exhibit poor performance on GUIOdyssey, which we attribute to a significant domain gap between cross-app and single-app tasks, resulting in substantial performance disparities. 
Supported by OmniParser's robust GUI grounding, most closed-source LVLMs substantially improve their cross-app performance, with Claude3.5-Sonnet achieving the best results.
In addition, OdysseyAgent surpasses the fine-tuned Qwen-VL, indicating that the proposed history resampler module enhances cross-app navigation. After incorporating semantic annotations during training, OdysseyAgent further improves its performance on all cross-app tasks, achieving $78.24$ and $88.15$ AMS in high-level and low-level instruction tasks, respectively, thereby demonstrating the effectiveness of our dataset.



\textbf{Out-of-domain Performance.}
We further assess the OdysseyAgent's generalization capability in unseen scenarios. 
As shown in \cref{tab:main_ood}, 
OdysseyAgent’s out-of-domain performance declines by $16.26$ and $5.92$ for high- and low-level instructions, respectively, compared to in-domain performance without semantic annotations.
With semantic annotations, these declines become $15.34$ and $6.95$. This suggests that high-level instructions are more challenging to generalize in cross-app tasks compared to low-level instructions.
Furthermore, incorporating semantic annotations during training improves performance in most scenarios, with especially notable gains on high-level instruction tasks, underscoring the value of semantic annotations for unseen domain.
Compared to in-domain performance, the performance gap between high- and low-level instructions is even larger in out-of-domain tasks. This implies that the model currently lacks sufficient reasoning and planning capabilities to effectively handle unseen high-level instruction tasks.

\subsection{The effect of different historical information.}\label{sec:exp-ablation}

We now conduct a detailed experiment to deeply explore the role of historical information components. Currently, two main types of historical information are used in GUI agents: historical actions and historical screenshots. Note that the Contextual Information included in the semantic annotations of GUIOdyssey serves as a summary of previous steps, providing a more comprehensive textual representation of historical information. Therefore, we also include it in our experiments. Detailed results are presented in \cref{tab:historical_comparison}. Comparing experiments (2)–(4) with the baseline experiment (1), we observe that all three types of historical information significantly improve model performance, with contextual information producing the most substantial enhancement: improving AMS by $9.17$ and SR by $240\%$ compared to the baseline.
A comparison between experiments (4) and (5) shows that using contextual information alone significantly improves out-of-domain performance compared to employing both actions and screenshots as historical input.
This suggests that summarizing and abstracting historical information can better help the model generalize to unseen GUI scenarios.
Additionally, experiment (6) shows that incorporating all types of historical information as input further enhances the model performance. 
We hypothesize that cross-app tasks inherently require more sophisticated memory mechanisms due to the dependencies and interactions between multiple apps. For example, as illustrated in \cref{fig:example} (Appendix \cref{sec:example}), completing a cross-app task—such as identifying properties of triangles from Chrome and subsequently recording them in Google Docs—requires effectively remembering and transferring key information across apps. This example highlights the critical role historical information plays, underscoring the importance of comprehensive historical context modeling for GUI agents in complex cross-app scenarios.


\textbf{More experiments.} To deepen our analysis using GUIOdyssey, we conduct additional experiments detailed in Appendix \cref{sec:more_exp}. These include investigations of various strategies for handling historical screenshots, different semantic annotation components, transferability across devices, different instruction granularities, and the relationship between cross-app and single-app tasks.

\section{Conclusion}

In this work, we address the limitations of existing GUI navigation agents for cross-app tasks by introducing GUIOdyssey, the first comprehensive cross-app mobile GUI navigation dataset enriched with semantic annotations. Leveraging this dataset, we develop OdysseyAgent, a multimodal cross-app navigation agent equipped with a history resampler module that efficiently processes historical image tokens to balance performance and inference speed.
We conduct extensive experiments with OdysseyAgent to evaluate our approach on both in-domain and out-of-domain scenarios. Our results further indicate that richer utilization of historical information can substantially enhance OdysseyAgent's performance.
We hope GUIOdyssey and OdysseyAgent can drive the research in the field of general GUI Agents.

\section*{Acknowledgments and Disclosure of Funding}\label{Sec:Ack}
We thank Zhouheng Yao, Zihao Zhao for their help in data collection. This paper is partially supported by the National Key R \& D Program of China No.2022ZD0160101 \& No.2022ZD0161000. 

{
    \small
    \bibliographystyle{ieeenat_fullname}
    \bibliography{main}

\begin{thebibliography}{69}
\providecommand{\natexlab}[1]{#1}
\providecommand{\url}[1]{\texttt{#1}}
\expandafter\ifx\csname urlstyle\endcsname\relax
  \providecommand{\doi}[1]{doi: #1}\else
  \providecommand{\doi}{doi: \begingroup \urlstyle{rm}\Url}\fi

\bibitem[Achiam et~al.(2023)Achiam, Adler, Agarwal, Ahmad, Akkaya, Aleman, Almeida, Altenschmidt, Altman, Anadkat, et~al.]{achiam2023gpt}
Josh Achiam, Steven Adler, Sandhini Agarwal, Lama Ahmad, Ilge Akkaya, Florencia~Leoni Aleman, Diogo Almeida, Janko Altenschmidt, Sam Altman, Shyamal Anadkat, et~al.
\newblock Gpt-4 technical report.
\newblock \emph{arXiv preprint arXiv:2303.08774}, 2023.

\bibitem[Anthropic(2023)]{Claude2023}
Anthropic.
\newblock Claude, 2023.
\newblock Accessed: 2023-04-18.

\bibitem[Baechler et~al.(2024)Baechler, Sunkara, Wang, Zubach, Mansoor, Etter, C{\u{a}}rbune, Lin, Chen, and Sharma]{baechler2024screenai}
Gilles Baechler, Srinivas Sunkara, Maria Wang, Fedir Zubach, Hassan Mansoor, Vincent Etter, Victor C{\u{a}}rbune, Jason Lin, Jindong Chen, and Abhanshu Sharma.
\newblock Screenai: A vision-language model for ui and infographics understanding.
\newblock \emph{arXiv preprint arXiv:2402.04615}, 2024.

\bibitem[Bai et~al.(2021)Bai, Zang, Xu, Sunkara, Rastogi, Chen, et~al.]{bai2021uibert}
Chongyang Bai, Xiaoxue Zang, Ying Xu, Srinivas Sunkara, Abhinav Rastogi, Jindong Chen, et~al.
\newblock Uibert: Learning generic multimodal representations for ui understanding.
\newblock \emph{arXiv preprint arXiv:2107.13731}, 2021.

\bibitem[Bai et~al.(2023)Bai, Bai, Yang, Wang, Tan, Wang, Lin, Zhou, and Zhou]{bai2023qwen}
Jinze Bai, Shuai Bai, Shusheng Yang, Shijie Wang, Sinan Tan, Peng Wang, Junyang Lin, Chang Zhou, and Jingren Zhou.
\newblock Qwen-vl: A frontier large vision-language model with versatile abilities.
\newblock \emph{arXiv preprint arXiv:2308.12966}, 2023.

\bibitem[Barkhuus and Polichar(2011)]{barkhuus2011empowerment}
Louise Barkhuus and Valerie~E Polichar.
\newblock Empowerment through seamfulness: smart phones in everyday life.
\newblock \emph{Personal and Ubiquitous Computing}, 15:\penalty0 629--639, 2011.

\bibitem[Biten et~al.(2019)Biten, Tito, Mafla, Gomez, Rusinol, Valveny, Jawahar, and Karatzas]{biten2019scene}
Ali~Furkan Biten, Ruben Tito, Andres Mafla, Lluis Gomez, Mar{\c{c}}al Rusinol, Ernest Valveny, CV Jawahar, and Dimosthenis Karatzas.
\newblock Scene text visual question answering.
\newblock In \emph{Proceedings of the IEEE/CVF international conference on computer vision}, pages 4291--4301, 2019.

\bibitem[Burns et~al.(2021)Burns, Arsan, Agrawal, Kumar, Saenko, and Plummer]{burns2021mobile}
Andrea Burns, Deniz Arsan, Sanjna Agrawal, Ranjitha Kumar, Kate Saenko, and Bryan~A Plummer.
\newblock Mobile app tasks with iterative feedback (motif): Addressing task feasibility in interactive visual environments.
\newblock \emph{arXiv preprint arXiv:2104.08560}, 2021.

\bibitem[Chai et~al.(2024)Chai, Huang, Niu, Xiao, Liu, Zhang, Gao, Ren, and Li]{chai2024amex}
Yuxiang Chai, Siyuan Huang, Yazhe Niu, Han Xiao, Liang Liu, Dingyu Zhang, Peng Gao, Shuai Ren, and Hongsheng Li.
\newblock Amex: Android multi-annotation expo dataset for mobile gui agents.
\newblock \emph{arXiv preprint arXiv:2407.17490}, 2024.

\bibitem[Chen et~al.(2024{\natexlab{a}})Chen, Huang, Wu, Tang, Chen, Bai, He, Wang, Zhou, Li, et~al.]{chen2024gui}
Dongping Chen, Yue Huang, Siyuan Wu, Jingyu Tang, Liuyi Chen, Yilin Bai, Zhigang He, Chenlong Wang, Huichi Zhou, Yiqiang Li, et~al.
\newblock Gui-world: A dataset for gui-oriented multimodal llm-based agents.
\newblock \emph{arXiv preprint arXiv:2406.10819}, 2024{\natexlab{a}}.

\bibitem[Chen et~al.(2024{\natexlab{b}})Chen, Cui, Hu, Qin, Fang, Zhao, Wang, Liu, Chen, Huo, et~al.]{chen2024guicourse}
Wentong Chen, Junbo Cui, Jinyi Hu, Yujia Qin, Junjie Fang, Yue Zhao, Chongyi Wang, Jun Liu, Guirong Chen, Yupeng Huo, et~al.
\newblock Guicourse: From general vision language models to versatile gui agents.
\newblock \emph{arXiv preprint arXiv:2406.11317}, 2024{\natexlab{b}}.

\bibitem[Chen et~al.(2023)Chen, Wu, Wang, Su, Chen, Xing, Zhong, Zhang, Zhu, Lu, Li, Luo, Lu, Qiao, and Dai]{chen2023internvl}
Zhe Chen, Jiannan Wu, Wenhai Wang, Weijie Su, Guo Chen, Sen Xing, Muyan Zhong, Qinglong Zhang, Xizhou Zhu, Lewei Lu, Bin Li, Ping Luo, Tong Lu, Yu Qiao, and Jifeng Dai.
\newblock Internvl: Scaling up vision foundation models and aligning for generic visual-linguistic tasks.
\newblock \emph{arXiv preprint arXiv:2312.14238}, 2023.

\bibitem[Chen et~al.(2024{\natexlab{c}})Chen, Wang, Tian, Ye, Gao, Cui, Tong, Hu, Luo, Ma, et~al.]{chen2024far}
Zhe Chen, Weiyun Wang, Hao Tian, Shenglong Ye, Zhangwei Gao, Erfei Cui, Wenwen Tong, Kongzhi Hu, Jiapeng Luo, Zheng Ma, et~al.
\newblock How far are we to gpt-4v? closing the gap to commercial multimodal models with open-source suites.
\newblock \emph{arXiv preprint arXiv:2404.16821}, 2024{\natexlab{c}}.

\bibitem[Cheng et~al.(2024)Cheng, Sun, Chu, Xu, Li, Zhang, and Wu]{cheng2024seeclick}
Kanzhi Cheng, Qiushi Sun, Yougang Chu, Fangzhi Xu, Yantao Li, Jianbing Zhang, and Zhiyong Wu.
\newblock Seeclick: Harnessing gui grounding for advanced visual gui agents.
\newblock \emph{arXiv preprint arXiv:2401.10935}, 2024.

\bibitem[Chi et~al.(2023)Chi, Yang, Liu, Han, Chen, Huang, Fu, and Yu]{zhang2023appagent}
Zhang Chi, Zhao Yang, Jiaxuan Liu, Yucheng Han, Xin Chen, Zebiao Huang, Bin Fu, and Gang Yu.
\newblock Appagent: Multimodal agents as smartphone users.
\newblock \emph{arXiv preprint arXiv:2312.13771}, 2023.

\bibitem[Chowdhery et~al.(2023)Chowdhery, Narang, Devlin, Bosma, Mishra, Roberts, Barham, Chung, Sutton, Gehrmann, et~al.]{chowdhery2023palm}
Aakanksha Chowdhery, Sharan Narang, Jacob Devlin, Maarten Bosma, Gaurav Mishra, Adam Roberts, Paul Barham, Hyung~Won Chung, Charles Sutton, Sebastian Gehrmann, et~al.
\newblock Palm: Scaling language modeling with pathways.
\newblock \emph{Journal of Machine Learning Research}, 24\penalty0 (240):\penalty0 1--113, 2023.

\bibitem[Deng et~al.(2024)Deng, Gu, Zheng, Chen, Stevens, Wang, Sun, and Su]{deng2024mind2web}
Xiang Deng, Yu Gu, Boyuan Zheng, Shijie Chen, Sam Stevens, Boshi Wang, Huan Sun, and Yu Su.
\newblock Mind2web: Towards a generalist agent for the web.
\newblock \emph{Advances in Neural Information Processing Systems}, 36, 2024.

\bibitem[Dubey et~al.(2024)Dubey, Jauhri, Pandey, Kadian, Al-Dahle, Letman, Mathur, Schelten, Yang, Fan, et~al.]{dubey2024llama3herdmodels}
Abhimanyu Dubey, Abhinav Jauhri, Abhinav Pandey, Abhishek Kadian, Ahmad Al-Dahle, Aiesha Letman, Akhil Mathur, Alan Schelten, Amy Yang, Angela Fan, et~al.
\newblock The llama 3 herd of models.
\newblock \emph{arXiv preprint arXiv:2407.21783}, 2024.

\bibitem[Furuta et~al.(2023)Furuta, Lee, Nachum, Matsuo, Faust, Gu, and Gur]{furuta2023multimodal}
Hiroki Furuta, Kuang-Huei Lee, Ofir Nachum, Yutaka Matsuo, Aleksandra Faust, Shixiang~Shane Gu, and Izzeddin Gur.
\newblock Multimodal web navigation with instruction-finetuned foundation models.
\newblock \emph{arXiv preprint arXiv:2305.11854}, 2023.

\bibitem[Gou et~al.(2024)Gou, Wang, Zheng, Xie, Chang, Shu, Sun, and Su]{gou2024uground}
Boyu Gou, Ruohan Wang, Boyuan Zheng, Yanan Xie, Cheng Chang, Yiheng Shu, Huan Sun, and Yu Su.
\newblock Navigating the digital world as humans do: Universal visual grounding for gui agents.
\newblock \emph{arXiv preprint arXiv:2410.05243}, 2024.

\bibitem[Gur et~al.(2023)Gur, Furuta, Huang, Safdari, Matsuo, Eck, and Faust]{gur2023real}
Izzeddin Gur, Hiroki Furuta, Austin Huang, Mustafa Safdari, Yutaka Matsuo, Douglas Eck, and Aleksandra Faust.
\newblock A real-world webagent with planning, long context understanding, and program synthesis.
\newblock \emph{arXiv preprint arXiv:2307.12856}, 2023.

\bibitem[He et~al.(2024)He, Yao, Ma, Yu, Dai, Zhang, Lan, and Yu]{he2024webvoyager}
Hongliang He, Wenlin Yao, Kaixin Ma, Wenhao Yu, Yong Dai, Hongming Zhang, Zhenzhong Lan, and Dong Yu.
\newblock Webvoyager: Building an end-to-end web agent with large multimodal models.
\newblock \emph{arXiv preprint arXiv:2401.13919}, 2024.

\bibitem[Hong et~al.(2023)Hong, Wang, Lv, Xu, Yu, Ji, Wang, Wang, Dong, Ding, et~al.]{hong2023cogagent}
Wenyi Hong, Weihan Wang, Qingsong Lv, Jiazheng Xu, Wenmeng Yu, Junhui Ji, Yan Wang, Zihan Wang, Yuxiao Dong, Ming Ding, et~al.
\newblock Cogagent: A visual language model for gui agents.
\newblock \emph{arXiv preprint arXiv:2312.08914}, 2023.

\bibitem[Kapoor et~al.(2024)Kapoor, Butala, Russak, Koh, Kamble, Alshikh, and Salakhutdinov]{kapoor2024omniact}
Raghav Kapoor, Yash~Parag Butala, Melisa Russak, Jing~Yu Koh, Kiran Kamble, Waseem Alshikh, and Ruslan Salakhutdinov.
\newblock Omniact: A dataset and benchmark for enabling multimodal generalist autonomous agents for desktop and web.
\newblock \emph{arXiv preprint arXiv:2402.17553}, 2024.

\bibitem[Koh et~al.(2024)Koh, Lo, Jang, Duvvur, Lim, Huang, Neubig, Zhou, Salakhutdinov, and Fried]{koh2024visualwebarena}
Jing~Yu Koh, Robert Lo, Lawrence Jang, Vikram Duvvur, Ming~Chong Lim, Po-Yu Huang, Graham Neubig, Shuyan Zhou, Ruslan Salakhutdinov, and Daniel Fried.
\newblock Visualwebarena: Evaluating multimodal agents on realistic visual web tasks.
\newblock \emph{arXiv preprint arXiv:2401.13649}, 2024.

\bibitem[Li et~al.(2024{\natexlab{a}})Li, Bishop, Li, Rawles, Campbell-Ajala, Tyamagundlu, and Riva]{li2024effects}
Wei Li, William Bishop, Alice Li, Chris Rawles, Folawiyo Campbell-Ajala, Divya Tyamagundlu, and Oriana Riva.
\newblock On the effects of data scale on computer control agents.
\newblock \emph{arXiv preprint arXiv:2406.03679}, 2024{\natexlab{a}}.

\bibitem[Li et~al.(2020)Li, He, Zhou, Zhang, and Baldridge]{li2020mapping}
Yang Li, Jiacong He, Xin Zhou, Yuan Zhang, and Jason Baldridge.
\newblock Mapping natural language instructions to mobile ui action sequences.
\newblock \emph{arXiv preprint arXiv:2005.03776}, 2020.

\bibitem[Li et~al.(2024{\natexlab{b}})Li, Zhang, Yang, Fu, Cheng, Chen, Chen, and Wei]{li2024appagent}
Yanda Li, Chi Zhang, Wanqi Yang, Bin Fu, Pei Cheng, Xin Chen, Ling Chen, and Yunchao Wei.
\newblock Appagent v2: Advanced agent for flexible mobile interactions.
\newblock \emph{arXiv preprint arXiv:2408.11824}, 2024{\natexlab{b}}.

\bibitem[Li et~al.(2024{\natexlab{c}})Li, You, Zhang, Feng, Agrawal, Li, Moorthy, Nichols, Yang, and Gan]{li2024ferret}
Zhangheng Li, Keen You, Haotian Zhang, Di Feng, Harsh Agrawal, Xiujun Li, Mohana Prasad~Sathya Moorthy, Jeff Nichols, Yinfei Yang, and Zhe Gan.
\newblock Ferret-ui 2: Mastering universal user interface understanding across platforms.
\newblock \emph{arXiv preprint arXiv:2410.18967}, 2024{\natexlab{c}}.

\bibitem[Liu et~al.(2024{\natexlab{a}})Liu, Li, Wu, and Lee]{liu2024visual}
Haotian Liu, Chunyuan Li, Qingyang Wu, and Yong~Jae Lee.
\newblock Visual instruction tuning.
\newblock \emph{Advances in neural information processing systems}, 36, 2024{\natexlab{a}}.

\bibitem[Liu et~al.(2024{\natexlab{b}})Liu, Song, Lin, Lam, Neubig, Li, and Yue]{liu2024visualwebbench}
Junpeng Liu, Yifan Song, Bill~Yuchen Lin, Wai Lam, Graham Neubig, Yuanzhi Li, and Xiang Yue.
\newblock Visualwebbench: How far have multimodal llms evolved in web page understanding and grounding?
\newblock \emph{arXiv preprint arXiv:2404.05955}, 2024{\natexlab{b}}.

\bibitem[L{\`u} et~al.(2024)L{\`u}, Kasner, and Reddy]{lu2024weblinx}
Xing~Han L{\`u}, Zden{\v{e}}k Kasner, and Siva Reddy.
\newblock Weblinx: Real-world website navigation with multi-turn dialogue.
\newblock \emph{arXiv preprint arXiv:2402.05930}, 2024.

\bibitem[Lu et~al.(2024)Lu, Yang, Shen, and Awadallah]{lu2024omniparserpurevisionbased}
Yadong Lu, Jianwei Yang, Yelong Shen, and Ahmed Awadallah.
\newblock Omniparser for pure vision based gui agent.
\newblock \emph{arXiv preprint arXiv:2408.00203}, 2024.

\bibitem[Mu et~al.(2024)Mu, Zhang, Hu, Wang, Ding, Jin, Wang, Dai, Qiao, and Luo]{mu2024embodiedgpt}
Yao Mu, Qinglong Zhang, Mengkang Hu, Wenhai Wang, Mingyu Ding, Jun Jin, Bin Wang, Jifeng Dai, Yu Qiao, and Ping Luo.
\newblock Embodiedgpt: Vision-language pre-training via embodied chain of thought.
\newblock \emph{Advances in Neural Information Processing Systems}, 36, 2024.

\bibitem[Nanavati et~al.(2023)Nanavati, Ranganeni, and Cakmak]{nanavati2023physically}
Amal Nanavati, Vinitha Ranganeni, and Maya Cakmak.
\newblock Physically assistive robots: A systematic review of mobile and manipulator robots that physically assist people with disabilities.
\newblock \emph{Annual Review of Control, Robotics, and Autonomous Systems}, 7, 2023.

\bibitem[OpenAI(2024)]{gpt4o}
OpenAI.
\newblock Gpt4o, 2024.

\bibitem[Ravi et~al.(2024)Ravi, Gabeur, Hu, Hu, Ryali, Ma, Khedr, R{\"a}dle, Rolland, Gustafson, Mintun, Pan, Alwala, Carion, Wu, Girshick, Doll{\'a}r, and Feichtenhofer]{ravi2024sam2}
Nikhila Ravi, Valentin Gabeur, Yuan-Ting Hu, Ronghang Hu, Chaitanya Ryali, Tengyu Ma, Haitham Khedr, Roman R{\"a}dle, Chloe Rolland, Laura Gustafson, Eric Mintun, Junting Pan, Kalyan~Vasudev Alwala, Nicolas Carion, Chao-Yuan Wu, Ross Girshick, Piotr Doll{\'a}r, and Christoph Feichtenhofer.
\newblock Sam 2: Segment anything in images and videos.
\newblock \emph{arXiv preprint arXiv:2408.00714}, 2024.

\bibitem[Rawles et~al.(2023)Rawles, Li, Rodriguez, Riva, and Lillicrap]{rawles2023android}
Christopher Rawles, Alice Li, Daniel Rodriguez, Oriana Riva, and Timothy Lillicrap.
\newblock Android in the wild: A large-scale dataset for android device control.
\newblock \emph{arXiv preprint arXiv:2307.10088}, 2023.

\bibitem[Rawles et~al.(2024)Rawles, Clinckemaillie, Chang, Waltz, Lau, Fair, Li, Bishop, Li, Campbell-Ajala, et~al.]{rawles2024androidworld}
Christopher Rawles, Sarah Clinckemaillie, Yifan Chang, Jonathan Waltz, Gabrielle Lau, Marybeth Fair, Alice Li, William Bishop, Wei Li, Folawiyo Campbell-Ajala, et~al.
\newblock Androidworld: A dynamic benchmarking environment for autonomous agents.
\newblock \emph{arXiv preprint arXiv:2405.14573}, 2024.

\bibitem[Roziere et~al.(2023)Roziere, Gehring, Gloeckle, Sootla, Gat, Tan, Adi, Liu, Remez, Rapin, et~al.]{roziere2023code}
Baptiste Roziere, Jonas Gehring, Fabian Gloeckle, Sten Sootla, Itai Gat, Xiaoqing~Ellen Tan, Yossi Adi, Jingyu Liu, Tal Remez, J{\'e}r{\'e}my Rapin, et~al.
\newblock Code llama: Open foundation models for code.
\newblock \emph{arXiv preprint arXiv:2308.12950}, 2023.

\bibitem[Schick et~al.(2024)Schick, Dwivedi-Yu, Dess{\`\i}, Raileanu, Lomeli, Hambro, Zettlemoyer, Cancedda, and Scialom]{schick2024toolformer}
Timo Schick, Jane Dwivedi-Yu, Roberto Dess{\`\i}, Roberta Raileanu, Maria Lomeli, Eric Hambro, Luke Zettlemoyer, Nicola Cancedda, and Thomas Scialom.
\newblock Toolformer: Language models can teach themselves to use tools.
\newblock \emph{Advances in Neural Information Processing Systems}, 36, 2024.

\bibitem[Shaw et~al.(2023)Shaw, Joshi, Cohan, Berant, Pasupat, Hu, Khandelwal, Lee, and Toutanova]{shaw2023pixels}
Peter Shaw, Mandar Joshi, James Cohan, Jonathan Berant, Panupong Pasupat, Hexiang Hu, Urvashi Khandelwal, Kenton Lee, and Kristina Toutanova.
\newblock From pixels to ui actions: Learning to follow instructions via graphical user interfaces.
\newblock In \emph{Advances in Neural Information Processing Systems}, 2023.

\bibitem[Shen et~al.(2024)Shen, Song, Tan, Li, Lu, and Zhuang]{shen2024hugginggpt}
Yongliang Shen, Kaitao Song, Xu Tan, Dongsheng Li, Weiming Lu, and Yueting Zhuang.
\newblock Hugginggpt: Solving ai tasks with chatgpt and its friends in hugging face.
\newblock \emph{Advances in Neural Information Processing Systems}, 36, 2024.

\bibitem[Shi et~al.(2017)Shi, Karpathy, Fan, Hernandez, and Liang]{pmlr-v70-shi17a}
Tianlin Shi, Andrej Karpathy, Linxi Fan, Jonathan Hernandez, and Percy Liang.
\newblock World of bits: An open-domain platform for web-based agents.
\newblock In \emph{Proceedings of the 34th International Conference on Machine Learning}, pages 3135--3144. PMLR, 2017.

\bibitem[Sun et~al.(2022)Sun, Chen, Chen, Dai, Zhu, and Yu]{sun2022meta}
Liangtai Sun, Xingyu Chen, Lu Chen, Tianle Dai, Zichen Zhu, and Kai Yu.
\newblock Meta-gui: towards multi-modal conversational agents on mobile gui.
\newblock \emph{arXiv preprint arXiv:2205.11029}, 2022.

\bibitem[Touvron et~al.(2023)Touvron, Lavril, Izacard, Martinet, Lachaux, Lacroix, Rozi{\`e}re, Goyal, Hambro, Azhar, et~al.]{touvron2023llama}
Hugo Touvron, Thibaut Lavril, Gautier Izacard, Xavier Martinet, Marie-Anne Lachaux, Timoth{\'e}e Lacroix, Baptiste Rozi{\`e}re, Naman Goyal, Eric Hambro, Faisal Azhar, et~al.
\newblock Llama: Open and efficient foundation language models.
\newblock \emph{arXiv preprint arXiv:2302.13971}, 2023.

\bibitem[Venkatesh et~al.(2022)Venkatesh, Talukdar, and Narayanan]{venkatesh2022ugif}
Sagar~Gubbi Venkatesh, Partha Talukdar, and Srini Narayanan.
\newblock Ugif: Ui grounded instruction following.
\newblock \emph{arXiv preprint arXiv:2211.07615}, 2022.

\bibitem[Wang et~al.(2024{\natexlab{a}})Wang, Xu, Jia, Zhang, Yan, Shen, Zhang, Huang, and Sang]{wang2024mobile2}
Junyang Wang, Haiyang Xu, Haitao Jia, Xi Zhang, Ming Yan, Weizhou Shen, Ji Zhang, Fei Huang, and Jitao Sang.
\newblock Mobile-agent-v2: Mobile device operation assistant with effective navigation via multi-agent collaboration.
\newblock \emph{arXiv preprint arXiv:2406.01014}, 2024{\natexlab{a}}.

\bibitem[Wang et~al.(2024{\natexlab{b}})Wang, Xu, Ye, Yan, Shen, Zhang, Huang, and Sang]{wang2024mobile}
Junyang Wang, Haiyang Xu, Jiabo Ye, Ming Yan, Weizhou Shen, Ji Zhang, Fei Huang, and Jitao Sang.
\newblock Mobile-agent: Autonomous multi-modal mobile device agent with visual perception.
\newblock \emph{arXiv preprint arXiv:2401.16158}, 2024{\natexlab{b}}.

\bibitem[Wang et~al.(2024{\natexlab{c}})Wang, Bai, Tan, Wang, Fan, Bai, Chen, Liu, Wang, Ge, Fan, Dang, Du, Ren, Men, Liu, Zhou, Zhou, and Lin]{Qwen2VL}
Peng Wang, Shuai Bai, Sinan Tan, Shijie Wang, Zhihao Fan, Jinze Bai, Keqin Chen, Xuejing Liu, Jialin Wang, Wenbin Ge, Yang Fan, Kai Dang, Mengfei Du, Xuancheng Ren, Rui Men, Dayiheng Liu, Chang Zhou, Jingren Zhou, and Junyang Lin.
\newblock Qwen2-vl: Enhancing vision-language model's perception of the world at any resolution.
\newblock \emph{arXiv preprint arXiv:2409.12191}, 2024{\natexlab{c}}.

\bibitem[Wang et~al.(2024{\natexlab{d}})Wang, Bhandary, Wang, and Moore]{wang2024using}
Zhiping~Paul Wang, Priyanka Bhandary, Yizhou Wang, and Jason~H Moore.
\newblock Using gpt-4 to write a scientific review article: a pilot evaluation study.
\newblock \emph{bioRxiv}, pages 2024--04, 2024{\natexlab{d}}.

\bibitem[Wu et~al.(2024{\natexlab{a}})Wu, Han, Ding, Weng, Liu, Yao, Yu, and Kong]{wu2024copilot}
Zhiyong Wu, Chengcheng Han, Zichen Ding, Zhenmin Weng, Zhoumianze Liu, Shunyu Yao, Tao Yu, and Lingpeng Kong.
\newblock Os-copilot: Towards generalist computer agents with self-improvement.
\newblock \emph{arXiv preprint arXiv:2402.07456}, 2024{\natexlab{a}}.

\bibitem[Wu et~al.(2024{\natexlab{b}})Wu, Wu, Xu, Wang, Sun, Jia, Cheng, Ding, Chen, Liang, et~al.]{wu2024atlas}
Zhiyong Wu, Zhenyu Wu, Fangzhi Xu, Yian Wang, Qiushi Sun, Chengyou Jia, Kanzhi Cheng, Zichen Ding, Liheng Chen, Paul~Pu Liang, et~al.
\newblock Os-atlas: A foundation action model for generalist gui agents.
\newblock \emph{arXiv preprint arXiv:2410.23218}, 2024{\natexlab{b}}.

\bibitem[Xie et~al.(2024)Xie, Zhang, Chen, Li, Zhao, Cao, Hua, Cheng, Shin, Lei, et~al.]{OSWorld}
Tianbao Xie, Danyang Zhang, Jixuan Chen, Xiaochuan Li, Siheng Zhao, Ruisheng Cao, Toh~Jing Hua, Zhoujun Cheng, Dongchan Shin, Fangyu Lei, et~al.
\newblock Osworld: Benchmarking multimodal agents for open-ended tasks in real computer environments.
\newblock \emph{arXiv preprint arXiv:2404.07972}, 2024.

\bibitem[Xing et~al.(2024)Xing, Zhang, Xue, Chen, Yang, and Xiao]{xing2024understanding}
Mingzhe Xing, Rongkai Zhang, Hui Xue, Qi Chen, Fan Yang, and Zhen Xiao.
\newblock Understanding the weakness of large language model agents within a complex android environment.
\newblock \emph{arXiv preprint arXiv:2402.06596}, 2024.

\bibitem[Xu et~al.(2023)Xu, Sun, Zheng, Geng, Zhao, Feng, Tao, and Jiang]{xu2023wizardlm}
Can Xu, Qingfeng Sun, Kai Zheng, Xiubo Geng, Pu Zhao, Jiazhan Feng, Chongyang Tao, and Daxin Jiang.
\newblock Wizardlm: Empowering large language models to follow complex instructions.
\newblock \emph{arXiv preprint arXiv:2304.12244}, 2023.

\bibitem[Yan et~al.(2023)Yan, Yang, Zhu, Lin, Li, Wang, Yang, Zhong, McAuley, Gao, et~al.]{yan2023gpt}
An Yan, Zhengyuan Yang, Wanrong Zhu, Kevin Lin, Linjie Li, Jianfeng Wang, Jianwei Yang, Yiwu Zhong, Julian McAuley, Jianfeng Gao, et~al.
\newblock Gpt-4v in wonderland: Large multimodal models for zero-shot smartphone gui navigation.
\newblock \emph{arXiv preprint arXiv:2311.07562}, 2023.

\bibitem[Yang et~al.(2023{\natexlab{a}})Yang, Zhang, Li, Zou, Li, and Gao]{yang2023set}
Jianwei Yang, Hao Zhang, Feng Li, Xueyan Zou, Chunyuan Li, and Jianfeng Gao.
\newblock Set-of-mark prompting unleashes extraordinary visual grounding in gpt-4v.
\newblock \emph{arXiv preprint arXiv:2310.11441}, 2023{\natexlab{a}}.

\bibitem[Yang et~al.(2023{\natexlab{b}})Yang, Li, Wang, Lin, Azarnasab, Ahmed, Liu, Liu, Zeng, and Wang]{yang2023mm}
Zhengyuan Yang, Linjie Li, Jianfeng Wang, Kevin Lin, Ehsan Azarnasab, Faisal Ahmed, Zicheng Liu, Ce Liu, Michael Zeng, and Lijuan Wang.
\newblock Mm-react: Prompting chatgpt for multimodal reasoning and action.
\newblock \emph{arXiv preprint arXiv:2303.11381}, 2023{\natexlab{b}}.

\bibitem[Yao et~al.(2022{\natexlab{a}})Yao, Chen, Yang, and Narasimhan]{yao2022webshop}
Shunyu Yao, Howard Chen, John Yang, and Karthik Narasimhan.
\newblock Webshop: Towards scalable real-world web interaction with grounded language agents.
\newblock \emph{Advances in Neural Information Processing Systems}, 35:\penalty0 20744--20757, 2022{\natexlab{a}}.

\bibitem[Yao et~al.(2022{\natexlab{b}})Yao, Zhao, Yu, Du, Shafran, Narasimhan, and Cao]{yao2022react}
Shunyu Yao, Jeffrey Zhao, Dian Yu, Nan Du, Izhak Shafran, Karthik Narasimhan, and Yuan Cao.
\newblock React: Synergizing reasoning and acting in language models.
\newblock \emph{arXiv preprint arXiv:2210.03629}, 2022{\natexlab{b}}.

\bibitem[Ying et~al.(2024)Ying, Meng, Wang, Li, Lin, Yang, Zhang, Zhang, Lin, Liu, et~al.]{ying2024mmt}
Kaining Ying, Fanqing Meng, Jin Wang, Zhiqian Li, Han Lin, Yue Yang, Hao Zhang, Wenbo Zhang, Yuqi Lin, Shuo Liu, et~al.
\newblock Mmt-bench: A comprehensive multimodal benchmark for evaluating large vision-language models towards multitask agi.
\newblock \emph{arXiv preprint arXiv:2404.16006}, 2024.

\bibitem[You et~al.(2025)You, Zhang, Schoop, Weers, Swearngin, Nichols, Yang, and Gan]{you2025ferret}
Keen You, Haotian Zhang, Eldon Schoop, Floris Weers, Amanda Swearngin, Jeffrey Nichols, Yinfei Yang, and Zhe Gan.
\newblock Ferret-ui: Grounded mobile ui understanding with multimodal llms.
\newblock In \emph{European Conference on Computer Vision}, pages 240--255. Springer, 2025.

\bibitem[Zhan and Zhang(2023)]{zhan2023you}
Zhuosheng Zhan and Aston Zhang.
\newblock You only look at screens: Multimodal chain-of-action agents.
\newblock \emph{arXiv preprint arXiv:2309.11436}, 2023.

\bibitem[Zhang et~al.(2024{\natexlab{a}})Zhang, Li, He, Zhang, Qiao, Qin, Ma, Kang, Lin, Rajmohan, et~al.]{zhang2024ufo}
Chaoyun Zhang, Liqun Li, Shilin He, Xu Zhang, Bo Qiao, Si Qin, Minghua Ma, Yu Kang, Qingwei Lin, Saravan Rajmohan, et~al.
\newblock Ufo: A ui-focused agent for windows os interaction.
\newblock \emph{arXiv preprint arXiv:2402.07939}, 2024{\natexlab{a}}.

\bibitem[Zhang et~al.(2024{\natexlab{b}})Zhang, Wu, Teng, Liao, Xu, Xiao, Wei, and Tang]{zhang2024android}
Jiwen Zhang, Jihao Wu, Yihua Teng, Minghui Liao, Nuo Xu, Xiao Xiao, Zhongyu Wei, and Duyu Tang.
\newblock Android in the zoo: Chain-of-action-thought for gui agents.
\newblock \emph{arXiv preprint arXiv:2403.02713}, 2024{\natexlab{b}}.

\bibitem[Zheng et~al.(2024)Zheng, Gou, Kil, Sun, and Su]{zheng2024seeact}
Boyuan Zheng, Boyu Gou, Jihyung Kil, Huan Sun, and Yu Su.
\newblock Gpt-4v(ision) is a generalist web agent, if grounded.
\newblock In \emph{Forty-first International Conference on Machine Learning}, 2024.

\bibitem[Zhou et~al.(2023)Zhou, Xu, Zhu, Zhou, Lo, Sridhar, Cheng, Bisk, Fried, Alon, et~al.]{zhou2023webarena}
Shuyan Zhou, Frank~F Xu, Hao Zhu, Xuhui Zhou, Robert Lo, Abishek Sridhar, Xianyi Cheng, Yonatan Bisk, Daniel Fried, Uri Alon, et~al.
\newblock Webarena: A realistic web environment for building autonomous agents.
\newblock \emph{arXiv preprint arXiv:2307.13854}, 2023.

\bibitem[Zhu et~al.(2023)Zhu, Chen, Tian, Tao, Su, Yang, Huang, Li, Lu, Wang, et~al.]{zhu2023ghost}
Xizhou Zhu, Yuntao Chen, Hao Tian, Chenxin Tao, Weijie Su, Chenyu Yang, Gao Huang, Bin Li, Lewei Lu, Xiaogang Wang, et~al.
\newblock Ghost in the minecraft: Generally capable agents for open-world enviroments via large language models with text-based knowledge and memory.
\newblock \emph{arXiv preprint arXiv:2305.17144}, 2023.

\end{thebibliography}
}

\clearpage
\setcounter{page}{1}
\maketitlesupplementary

\begin{table*}[]
    \centering
    \caption{The argument and functionality of different actions in GUIOdyssey. `pos1' and `pos2' denote the position $(x,y)$.}
    \label{tab:action-space}
    \begin{tabular}{c|c | c}
    \toprule
        Action & Argument & Functionality \\
        \midrule
        \texttt{CLICK} & [pos1] & click the on-screen position\\
        \midrule
        \texttt{LONG PRESS} & [pos1] & press the screen for a long time to copy texts or download images  \\
        \midrule
        \texttt{SCROLL} & [pos1, pos2] & scroll the screen from position 1 to position 2 \\
        \midrule
        \texttt{TYPE} & text & type text with keyboard \\
        \midrule
        \texttt{COMPLETE} & - & the sign that the instruction has been completed \\
        \midrule
        \texttt{IMPOSSIBLE} & - & the sign that the instruction  cannot be completed \\
        \midrule
        \texttt{HOME} & - & go to the home screen \\
        \midrule
        \texttt{BACK} & - & go to the previous screen\\
        \midrule
        \texttt{RECENT} & - & go to the previous App \\
        \bottomrule
    \end{tabular}
\end{table*}

\section{Ethical Discussion}\label{sec:ethical}
\

\textbf{Privacy.} We use temporary accounts and virtual usernames to register various apps and ensure no personal information is entered. The dataset contains no authentic personal information.

\textbf{Ethical Consent in Data Collection.} A formal consent process is implemented, wherein participants explicitly agree to the inclusion of their human-annotated data in the dataset. All data are collected with informed consent and in full compliance with ethical guidelines.

\textbf{Security Concerns.} The development of intelligent agents trained on datasets like this offers significant potential for automating tasks and enhancing accessibility. However, it also raises important ethical and security concerns. Sensitive operations, such as financial transactions or privacy management, pose vulnerabilities without robust safeguards. Additionally, malicious actors could exploit these agents to bypass security protocols or manipulate applications for unethical purposes.
To mitigate these risks, it is crucial to implement secure model designs, privacy-preserving techniques, and establish clear ethical guidelines. Addressing these challenges will help ensure the responsible deployment of such technology while maximizing its societal benefits.

\section{Details of GUIOdyssey}\label{sec:apx-detail-gui-odyssey}

\subsection{Description of Task Categories}\label{sec:desc-gui-odyssey}

The specific details of the six task categories are as follows:

\textbf{General Tool.} This category encompasses tasks that involve navigating through system-wide operations such as managing system settings or notifications for apps. An instruction example of a general tool task is ``Adjust the notification settings for the YouTube app on your phone using Settings, then proceed to open YouTube".

\textbf{Information Management.} Information management tasks involve searching for information and recording it for future use. This might include looking up information on search engines, reading articles on news apps, checking facts on educational or reference apps, and then saving or organizing this information in note-taking apps.

\textbf{Web Shopping.} Shopping tasks encompass a range of activities related to purchasing products online. Users may start by searching for a product on one app, comparing prices on different e-commerce platforms, checking reviews and ratings on review apps or websites, and finally making a purchase. 

\textbf{Media Entertainment.} Media entertainment tasks are about activities involving video and music streaming apps. Users may browse for new content on video platforms like YouTube or Netflix, stream music on services like Spotify or Apple Music, and switch between different media apps to manage playlists or download content.

\textbf{Social Sharing.} This task involves activities where users share content across different social media platforms. This could include taking photos or videos with the camera app, editing them using a photo or video editing app, and then sharing them on multiple social media platforms like Instagram, Facebook, Twitter, or TikTok. 

\textbf{Multi-Apps.} Multiple-app tasks involve more complex operations that require three or more apps to complete. For example, cooking food with an online recipe might involve finding the recipe of the food, recording the recipe to a note-taking app, and buying the ingredients online(\cref{fig:single-cross-app}).

\subsection{Action Set}\label{sec:app-action-set}
Our recording system utilizes Android Studio to simulate GUI navigation and virtualize various devices. We use the Android Debug Bridge (ADB) to retrieve device information and status, such as the coordinates of click events, and to monitor a wide range of functional keys.
The details of the action set in our Android emulator are presented in \cref{tab:action-space}.

\subsection{Fine-grained Episode Annotation Generation}\label{sec:semantic_detail}
Fine-grained episode annotations consist of two components: low-level instructions and semantic annotations.
Examples of the fine-grained annotations can be found in \cref{fig:example_anno}.

\begin{figure*}[h!]
    \centering
    \includegraphics[width=1.0\textwidth]{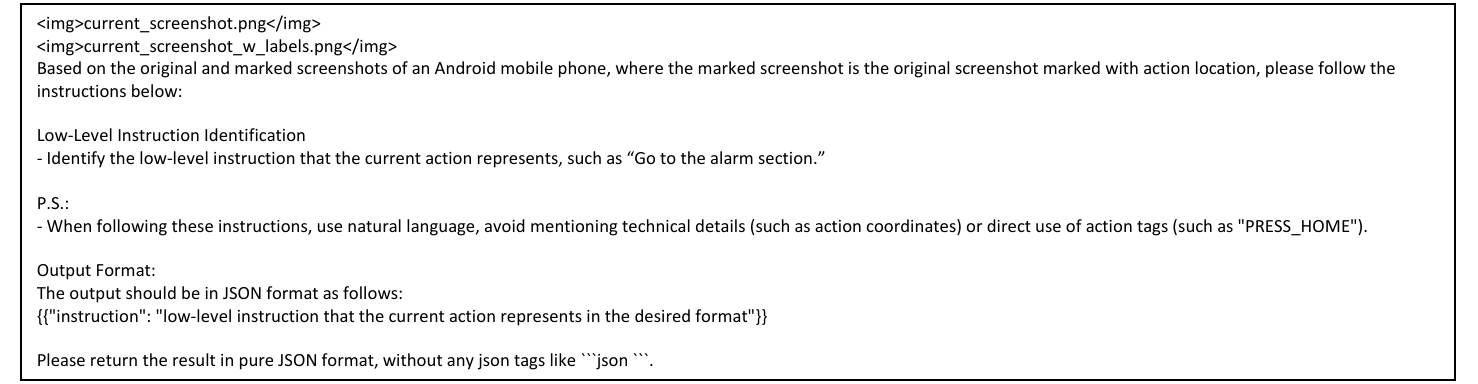}
    \caption{Prompts for generating low-level instruction.}
    \label{fig:prompt_lli}
\end{figure*}

\textbf{Low-Level Instruction.}
For each step within an episode, we provide GPT-4o with the high-level instruction corresponding to the episode, along with the action and screenshot associated with the current step. Additionally, for actions such as \texttt{CLICK} and \texttt{LONG PRESS}, we supply an additional image featuring a bounding box to indicate the click coordinates. All images are configured with the fidelity parameter set to `high'. The prompt utilized is provided in \cref{fig:prompt_lli}.

\begin{figure*}[h!]
    \centering
    \includegraphics[width=1.0\textwidth]{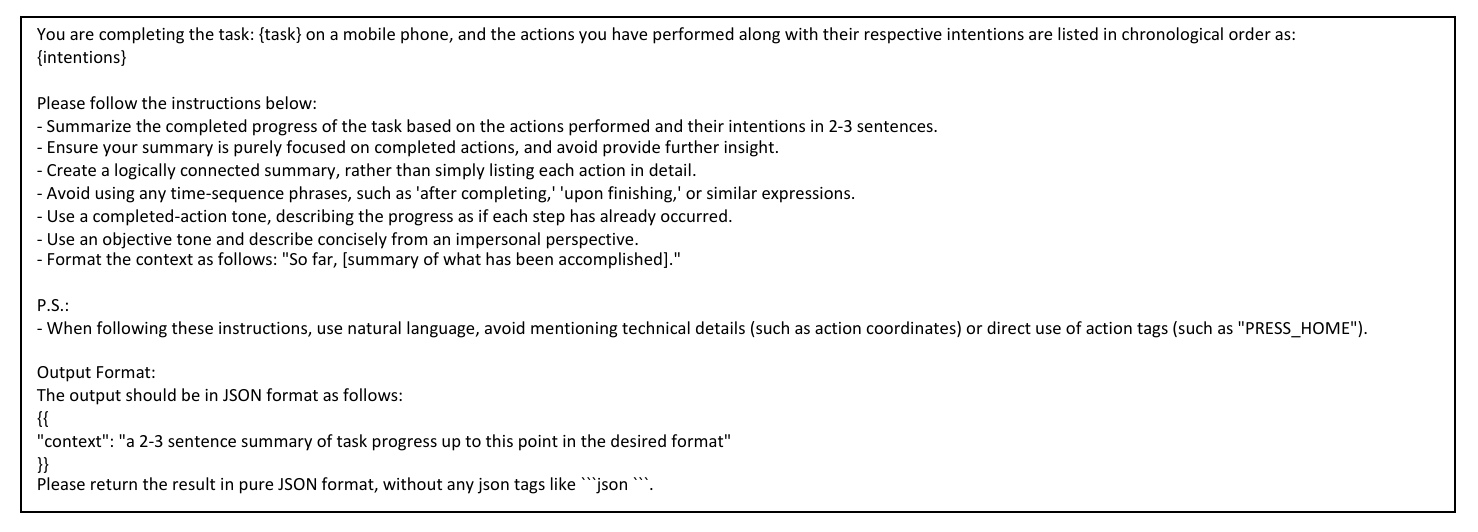}
    \caption{Prompts for generating contextual information.}
    \label{fig:prompt_ci}
\end{figure*}

\begin{figure*}[h]
    \centering
    \includegraphics[width=1.0\textwidth]{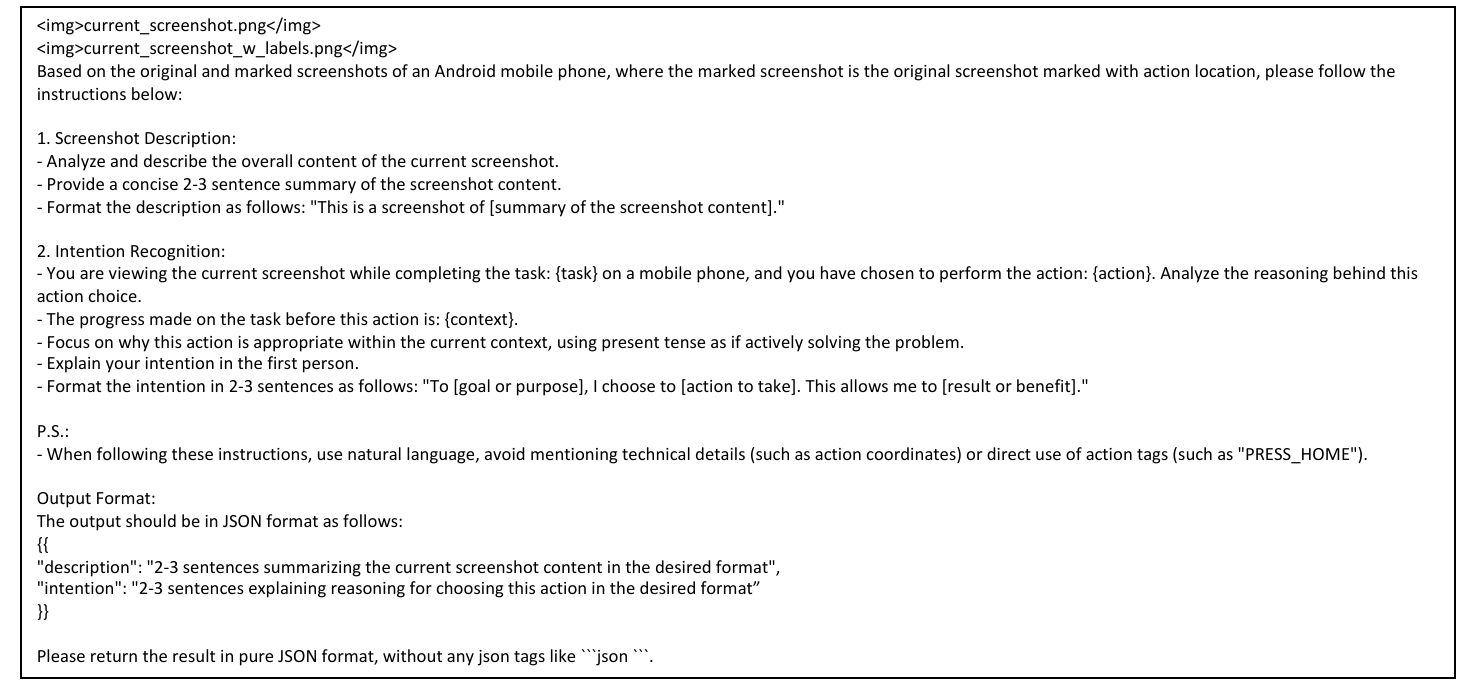}
    \caption{Prompts for generating screen description and decision rationale.}
    \label{fig:prompt_sa}
\end{figure*}

\textbf{Semantic Annotation.}
We use GPT-4o to generate semantic annotations in an alternating and iterative manner, following the sequential order of steps within each episode. Specifically, the process begins by providing the current episode's high-level instruction along with the actions and decision rationale from previous steps, prompting GPT-4o to generate the contextual information for the current step. Subsequently, using the generated contextual information, the high-level instruction, the screenshot image, and the action corresponding to the current step, GPT-4o is prompted step-by-step to generate the screen description and decision rationale for the current step. This iterative process continues until all semantic annotations for each step within the episode are completed in sequence.
Similarly, for actions such as \texttt{CLICK} and \texttt{LONG PRESS}, we supply an additional image with a bounding box indicating the click coordinates. All images are configured with the fidelity parameter set to `high' to ensure precision. The prompts used for generating these annotations are provided in \cref{fig:prompt_ci} and \cref{fig:prompt_sa}.

\subsection{Examples}\label{sec:example}
\begin{figure*}[h!]
    \centering
    \includegraphics[width=0.95\textwidth]{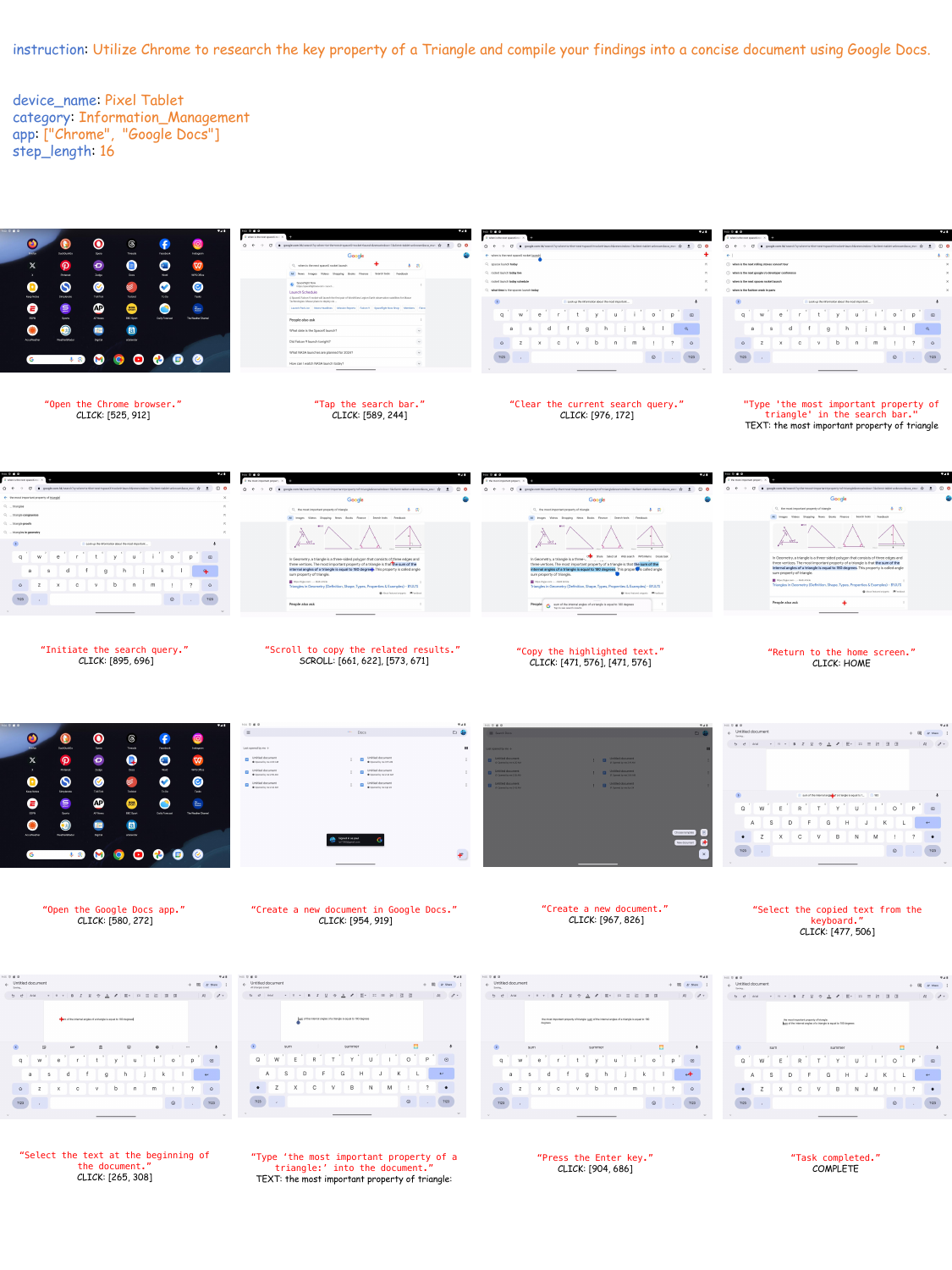}
    \caption{An example of episodes in our GUIOdyssey.}
    \label{fig:example}
\end{figure*}

\begin{figure*}[h!]
    \centering
    \includegraphics[width=0.7\textwidth]{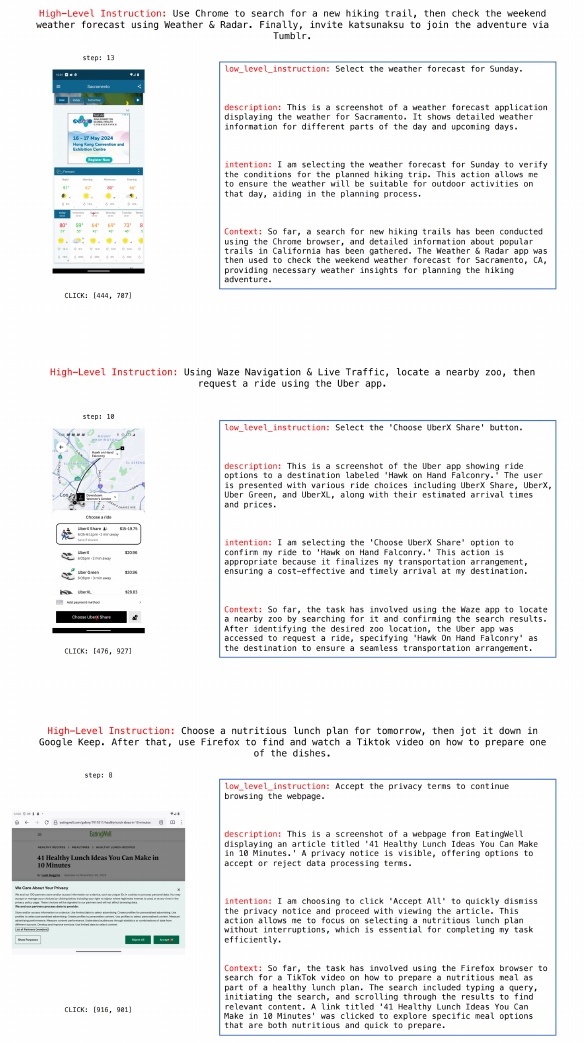}
    \caption{Examples of fine-grained annotations in GUIOdyssey.}
    \label{fig:example_anno}
\end{figure*}

\begin{figure*}[h!]
    \centering
    \includegraphics[width=0.7\textwidth]{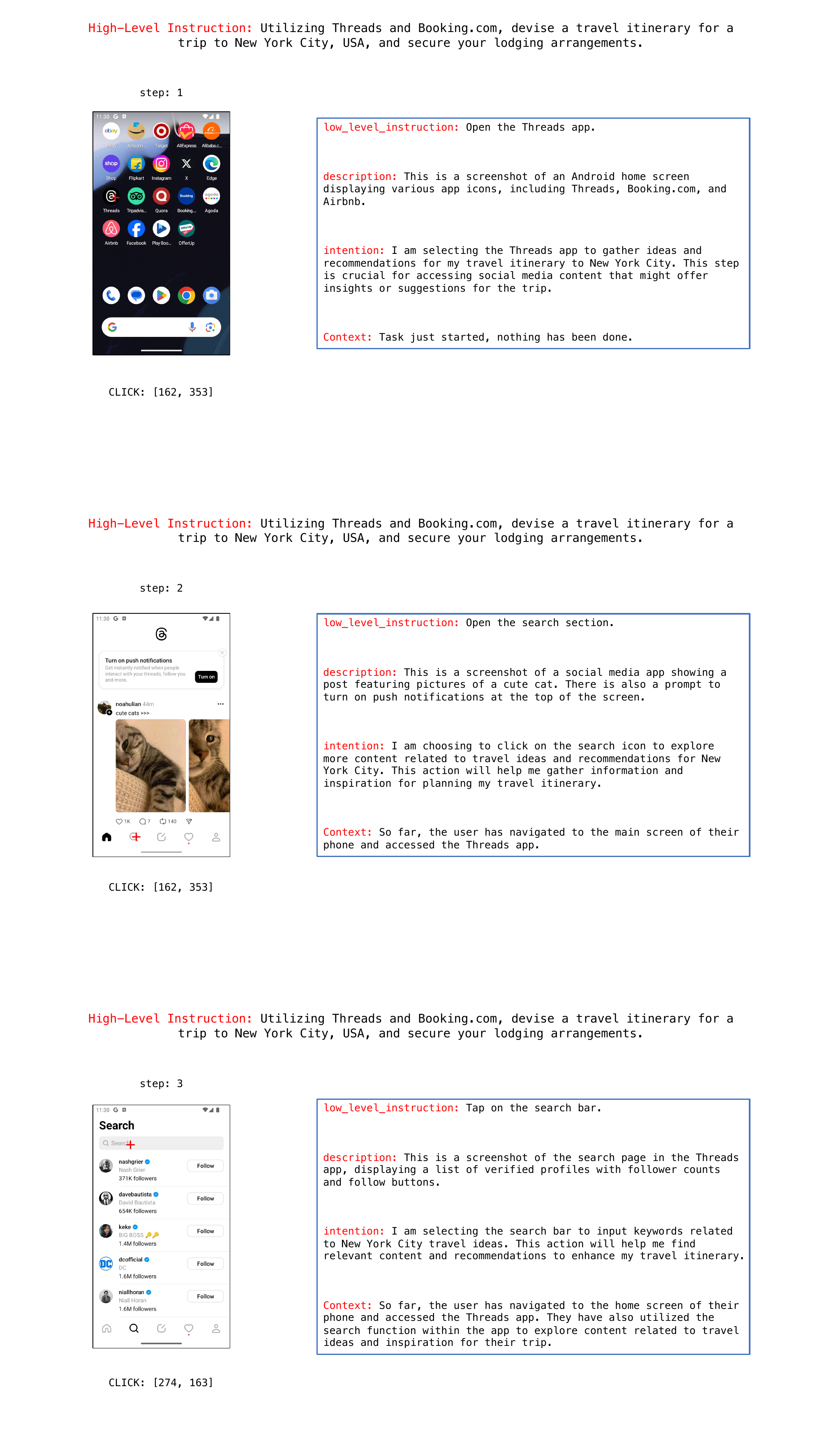}
    \caption{Example of an annotation for an unsuccessful task, ending with the \texttt{IMPOSSIBLE} action.}
    \label{fig:example_not_com1}
\end{figure*}

\begin{figure*}[h!]
    \centering
    \includegraphics[width=0.7\textwidth]{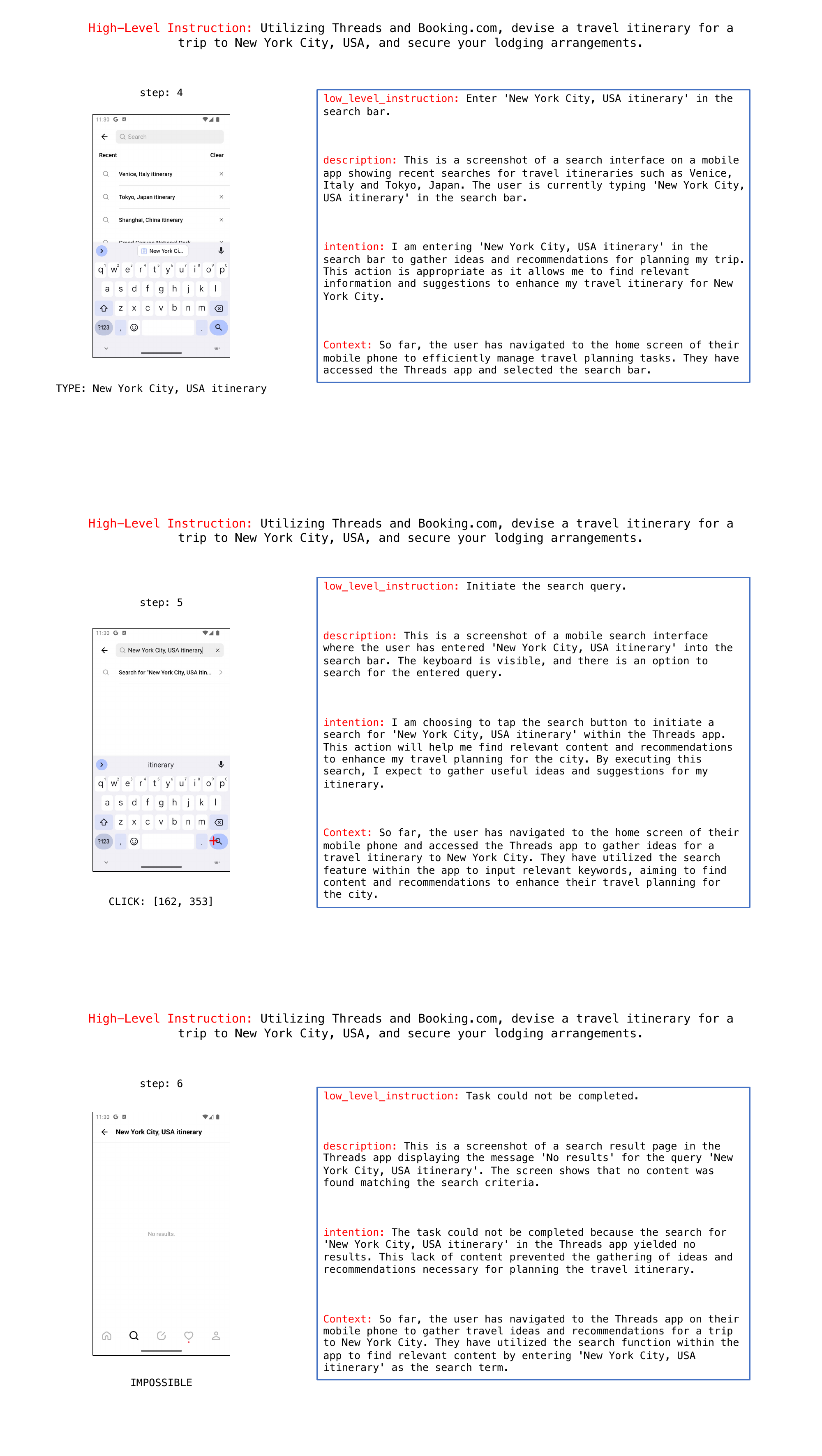}
    \caption{Example of an annotation for an unsuccessful task, ending with the \texttt{IMPOSSIBLE} action.}
    \label{fig:example_not_com2}
\end{figure*}

An example of episodes in our GUIOdyssey is shown in \cref{fig:example}, while examples of semantic annotations can be found in \cref{fig:example_anno}. An example of an annotation for a task that could not be successfully completed and ends with the \texttt{IMPOSSIBLE} action can be found in \cref{fig:example_not_com1} and \cref{fig:example_not_com2}.

\begin{figure*}[h!]
    \centering
    \includegraphics[width=0.9\textwidth]{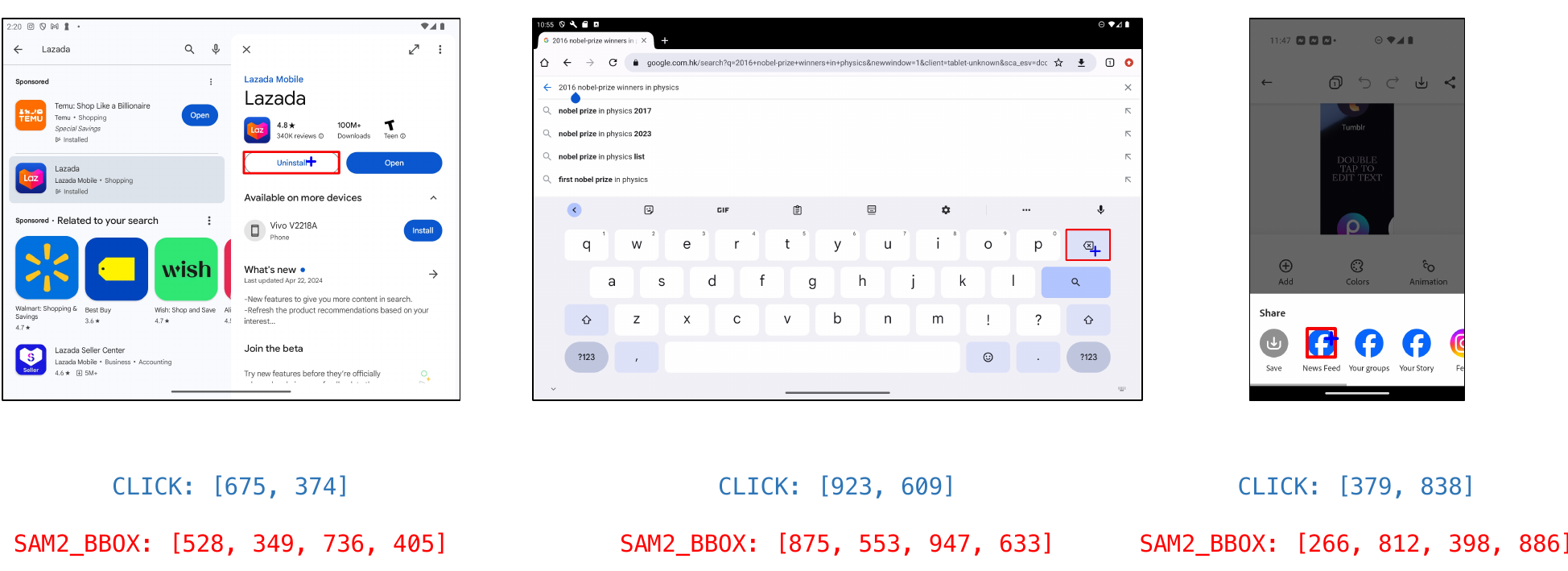}
    \caption{Examples of bounding boxes for UI elements segmented by SAM2. The actual click locations are indicated by blue `+' symbols, while the red rectangles outline the bounding boxes obtained from the SAM2.}
    \label{fig:example_sam2}
\end{figure*}

As mentioned in \cref{sec:exp-setup}, we use SAM2 \cite{ravi2024sam2} to assist in evaluating whether the model's output actions are correct. \cref{fig:example_sam2} provides examples of bounding boxes for clicked elements obtained through SAM2 segmentation.

\subsection{Data Format}

Each field of annotation is as follows. 

\textbf{episode\_id:} the unique identifier of this episode.

\textbf{device\_info:} the detailed information of the virtual device from which the episode was collected, including the device model, screen resolution, and other device-related details.

\textbf{task\_info:} the detailed information of the task from which the episode was collected, including the task category, the app used, the high-level instruction, and other task-related details.

\textbf{step\_length:} the total number of steps in this episode.

\textbf{steps:} a list of steps in this episode. Each step in the list includes the file path of the screenshot, executed action and its corresponding parameters (\eg, the coordinates for a click action), the low-level instruction, the semantic annotation, the bounding box obtained from SAM2 segmentation, and additional recorded information such as the overall scroll trajectory for scroll actions and annotator notes.

\begin{table*}[t!]
\centering
\caption{The impact of different semantic annotations on OdysseyAgent across four different splits. We use high-level instructions for both training and evaluation. Performance is assessed using AMS and SR as metrics. SD, CI, and DR denote screen description, contextual information, and decision rationale, respectively.}
\label{tab:semantic_comparison}
\scalebox{1}{
\begin{tabular}{c|ccc| cc | cc | cc | cc|cc}
\toprule

 & \multicolumn{3}{c|}{\textbf{Semantic Annotation}} & \multicolumn{2}{c|}{\textbf{Test-Random}} & \multicolumn{2}{c|}{\textbf{Test-Task}}  & \multicolumn{2}{c|}{\textbf{Test-Device}} & \multicolumn{2}{c|}{\textbf{Test-App}} & \multicolumn{2}{c}{\textbf{Overall}} \\ 
 & SD & CI & DR & AMS & SR & AMS & SR & AMS & SR & AMS & SR & AMS & SR \\
\midrule
(1) & \ding{55} & \ding{55} & \ding{55} & 75.79 & 9.38 & 54.36 & 0.09 & 61.20 & 1.88 & 63.03 & 7.70 & 63.60 & 4.76 \\
\midrule
(2) & \ding{51} & \ding{55} & \ding{55} & 75.18 & 8.94 & 54.06 & 0.00 & 64.41 & 2.03 & 64.91 & 8.47 & 64.64 & 4.86 \\
(3) & \ding{55} & \ding{51} & \ding{55} & 75.42 & 10.04 & 55.71 & 0.00 & 62.52 & 3.19 & 64.24 & 5.30 & 64.47 & 4.63\\ 
(4) & \ding{55} & \ding{55} & \ding{51} & 77.71 & 11.44 & 55.60 & \textbf{0.26} & 65.88 & 4.63 & 65.74 & 7.96 & 66.23 & 6.07 \\
\midrule
(5) & \ding{55} & \ding{51} & \ding{51} & 77.23 & 11.16 & 56.93 & 0.18 & 63.87 & 2.24 & 66.32 & 7.87 & 66.09 & 5.36 \\
(6) & \ding{51} & \ding{55} & \ding{51} & 77.24 & 10.88 & \textbf{57.15} & 0.00 & 63.55 & 2.17 & \textbf{67.04} & \textbf{9.67} & 66.24 & 5.68 \\
(7) & \ding{51} & \ding{51} & \ding{55} & 76.58 & 10.14 & 57.13 & \textbf{0.26}& 64.48 & 3.91 & 66.27 & 7.96 & 66.11 & 5.57 \\
\midrule
(8) & \ding{51} & \ding{51} & \ding{51} & \textbf{78.24} & \textbf{11.62} & 56.19 & \textbf{0.26} & \textbf{66.63} & \textbf{5.07} & 65.89 & 8.81 & \textbf{66.74} & \textbf{6.44} \\
\bottomrule
\end{tabular}
}
\end{table*}

\section{Experiment Details}\label{sec:exp-detailed}

\subsection{Detailed description of four different setups.}\label{sec:split-desc}

The following details the four different setups in GUIOdyssey.

\textbf{i) Train-Random \& Test-Random.} We randomly partitioned all the episodes in the dataset into training and testing sets using a ratio of $80\%$ to $20\%$ as the standard approach to divide the dataset. It can assess the in-domain performance of OdysseyAgent.

\textbf{ii) Train-Task \& Test-Task.} In this setup, We proportionally sampled meta-tasks from six categories, maintaining approximately a $6:1$ ratio for the training and test sets. The tasks in the test set differ significantly from those in the training set. This partitioning method allows for a robust assessment of an agent's generalization capabilities across diverse tasks. 

\textbf{iii) Train-Device \& Test-Device.} To evaluate an agent's generalizability across different and unseen devices, we selected episodes annotated on the Tablet, which differs significantly from other devices, as the test set. We obtained $1,381$ episodes as the test set and $6,953$ episodes as the training set.

\textbf{iv) Train-App \& Test-App.} This split is aimed at evaluating the agent's performance on unseen Apps and App combinations. First, we calculated the frequency of app usage in the dataset and categorized the apps into $25$ classes (\eg, Video, Music) based on their characteristics. Then, we selected a few apps with the lowest occurrence from each class to form the test app set. Subsequently, we partitioned the episodes that utilized the app in the test app set into the Test-App set, maintaining an approximately $85\%$ to $15\%$ ratio between the training set and the test set.

\subsection{Training Details.} \label{sec:training_datail}
To train OdysseyAgent, we employ the AdamW optimizer with a learning rate of $2e$$-5$ and utilize a cosine learning rate schedule. We set $\beta_1$ and $\beta_2$ to $0.9$ and $0.95$, respectively, and use a weight decay of $0.1$. Additionally, we utilize a global batch size of $128$ and implement DeepSpeed ZERO2-style data parallelism. 
During training, OdysseyAgent treats each action step as an individual training sample. The input consists of the task instruction, the current screenshot, and the previous $4$ actions and screenshots (\emph{i.e.}, $\delta = 4$), while the output corresponds to the action for the current step.
By default, OdysseyAgent is trained separately on Train-Random/Task/Device/App for one epoch, excluding the semantic annotation component. When training includes semantic annotations, these annotations are converted into single-turn QA pairs, which serve as additional training samples (\emph{i.e.}, semantic annotations are introduced only during training-time). Any training configuration that incorporates semantic annotations is explicitly noted. The entire training process requires approximately $32$ A100 hours to complete.

\subsection{Prompt for Evaluation.}\label{sec:evaluation-prompt}

\begin{figure*}[h]
    \centering
    \includegraphics[width=1.0\textwidth]{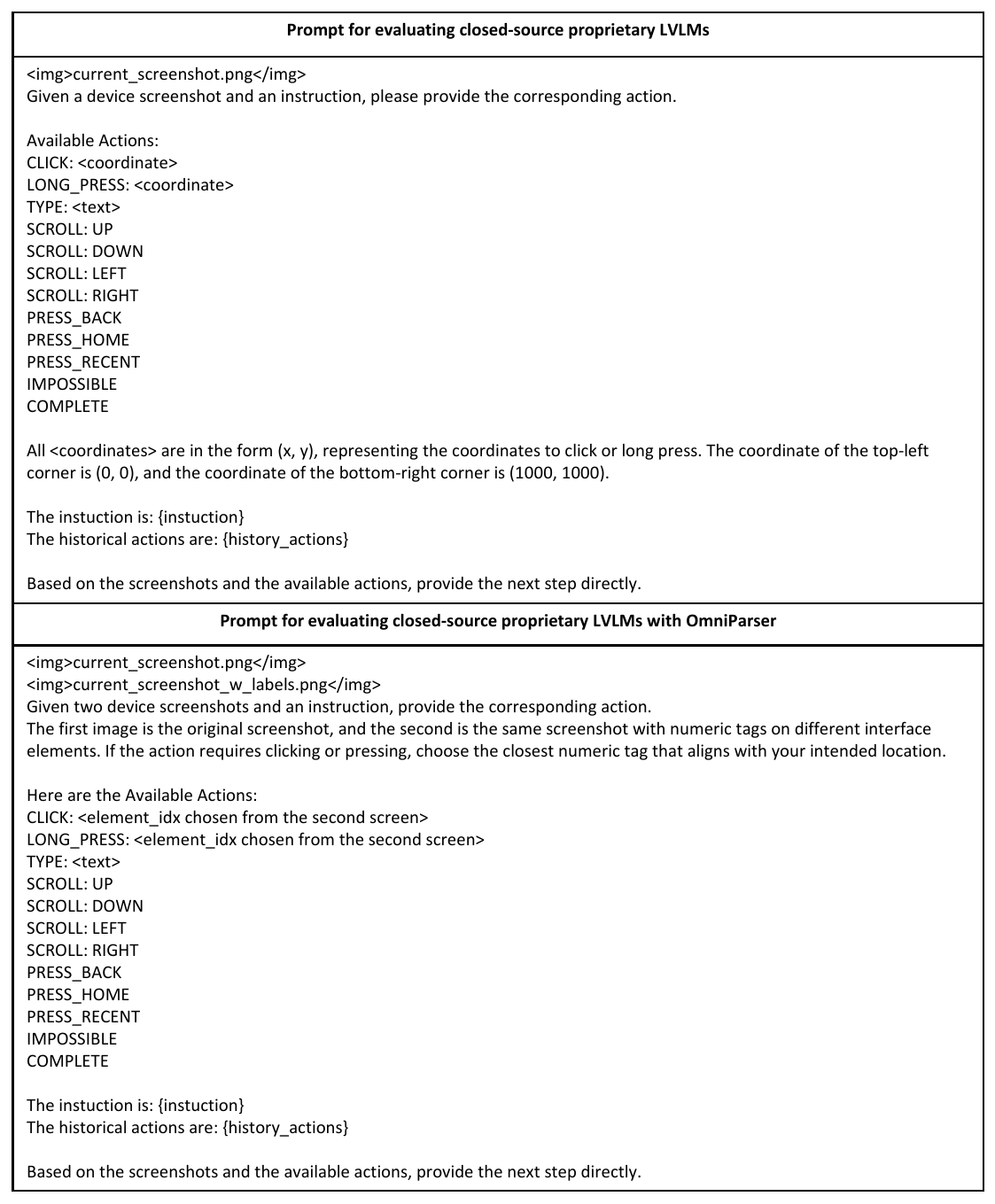}
    \caption{The prompt for evaluating closed-source proprietary Large Vision Language Models (LVLMs).}
    \label{fig:prompt-for-other}
\end{figure*}

We utilize the prompt shown in \cref{fig:prompt-for-other} to evaluate the performance of GPT-4V, GPT-4o, Claude3.5-sonnet, and InternVL2-Pro. For SphAgent and CogAgent, we tested them following their officially recommended methods~\cite{chai2024amex, hong2023cogagent}.

\section{More Experiments}\label{sec:more_exp}

\subsection{History Resampler vs. Multi-Image Training.}\label{sec:strategy_vs}
We evaluate different approaches for processing historical screenshot images. Qwen-VL supports multi-image input by interleaving image and text tokens, but this leads to a high token overhead (\eg, $1024$ tokens for four historical steps). 
Our history resampler compresses this to $256$ tokens, greatly improving efficiency. As shown in \cref{tab:strategy}, both approaches achieve comparable performance, but the history resampler significantly enhances training and inference efficiency.

\begin{table}[!h]
\centering
\caption{The average AMS for HL and LL instructions across 4 splits, along with the number of historical screenshot tokens, inference metrics (Time to First Token (TTFT) and Tokens per Second (TPS)), and training GPU hours.}
\label{tab:strategy}
\scalebox{0.65}{
\begin{tabular}{l|cc|c|cc|c}
\toprule
strategy  & HL & LL & Token Count & TTFT $\downarrow$ & TPS $\uparrow$ & GPU Hours\\
\midrule
history resampler   & 63.60 & 82.44 & 256 & 0.71 & 20.27 & 32 \\
multi-image   & 65.04 & 82.34 & 1024 & 0.98 & 17.05 & 48 \\
\bottomrule
\end{tabular}
}
\end{table}


\subsection{The effect of different semantic annotations.} 
We assess the impact of different semantic annotations in GUIOdyssey (\emph{i.e.}, screen description, contextual information and decision rationale) on model performance in both in-domain and out-of-domain settings. The results are presented in \cref{tab:semantic_comparison}.
A comparison of experiments (1)–(4) shows that all three components contribute positively, but engaging in detailed reasoning before making decisions is more important than understanding current screen information or summarizing historical processes in cross-app tasks. Experiments (5)–(8) further indicate that using two or more types of semantic annotations generally outperforms using a single annotation type. Specifically, using all semantic annotations yields the best results and improves AMS by $3.14$ and SR by $35\%$ compared to training without any semantic annotations. These findings suggest that teaching the model to understand the reasoning behind each action—similar to how humans observe, understand, review completed steps, and reason thoroughly before deciding—can be beneficial for improving performance in both in-domain and out-of-domain cross-app tasks.


\subsection{Transferability of instructions at different levels of granularity.}
As shown in \cref{tab:instruction_comparison}, models trained on high-level instructions exhibit significantly better transferability across different levels of instruction granularity compared to those trained on low-level instructions. Furthermore, training on both instruction granularities outperforms training on a single granularity, a phenomenon similar to what has been observed in single-app tasks \cite{li2024effects}.

\subsection{Transferability across different devices.}  
We utilize our GUIOdyssey dataset to conduct additional experiments to evaluate the generalization capabilities of OdysseyAgent beyond the initial experimental setup. we test the OdysseyAgent's adaptability by using data from one device as the test set while training on data from the remaining five devices.
The results of these experiments are presented in the \cref{tab:device_comparison}, demonstrating the model's performance across different devices. The model exhibits the weakest transferability on tablet devices, which we attribute to the significant differences in user interface layouts between tablets and smartphones. Furthermore, the model's transferability on small phones and foldable devices is also suboptimal. We surmise that the disparity in screen resolution compared to other phone models may contribute to this underperformance.


\begin{table}[h!]
    \centering
    \caption{The results for OdysseyAgent trained and tested on Train-Random/Test-Random with both high-level and low-level instructions are presented, with AMS as the evaluation metric. HL and LL denote high-level and low-level instructions, respectively.}
    \label{tab:instruction_comparison}
    \scalebox{0.9}{
    \begin{tabular}{c|ccc}
        \toprule
        \multirow{2}*{\textbf{Testing Instructions}}  & \multicolumn{3}{c}{\textbf{Training Instructions}} \\
         & HL & LL & HL + LL \\
        \midrule
        HL & 75.79 & 29.39 & 78.96 \\
        LL & 71.26 & 86.88 & 88.84 \\
\bottomrule
    \end{tabular}
}
\end{table}

\begin{table}[h!]
    \centering
    \caption{Performance Evaluation of OdysseyAgent Across Different Devices. Each Device serves as a test set while the remaining five devices are used as training sets.}
    \label{tab:device_comparison}
    \scalebox{1}{
    \begin{tabular}{l|l |cc}
        \toprule
        \textbf{Evaluation Device} & \textbf{Resolution} & \textbf{AMS} & \textbf{SR}  \\
        \midrule
        Pixel 7 Pro & $1,440 \times 3,120$ & 75.91 & 7.44  \\
        Pixel 8 Pro & $1,344 \times 2,992$ & 74.67 & 6.05 \\
        Small Phone & $720 \times 1,280$  &71.68 & 3.77 \\
        Medium Phone & $1,080 \times 2,400$ & 73.05 & 5.45  \\
        Pixel Fold  & $2,208 \times 1,840$ & 67.67 & 4.48  \\
        Pixel Tablet & $2,560 \times 1,600$ & 61.20 & 1.88  \\
        \bottomrule
    \end{tabular}
}
\end{table}

\subsection{Whether cross-App tasks benefit single-App tasks.}

We further investigate whether cross-app tasks benefit single-app performance by evaluating the impact of different training data compositions under controlled conditions. Specifically, we randomly sample 50k training samples each from GUIOdyssey, AITW, and AndroidControl (denoted as Ody50k, AITW50k, and AC50k, respectively) and evaluate their performance on AndroidControl, which provides both in-domain and out-of-domain scenarios. As shown in \cref{tab:androidcontrol_comparison}, we find that incorporating cross-app data from GUIOdyssey consistently enhances performance in most single-app scenarios, whereas adding AITW data surprisingly yields limited improvements or even performance degradation. This suggests that the more complex cross-app tasks in GUIOdyssey can benefit single-app tasks.

\begin{table}[h!]
    \centering
    \caption{Effectiveness of Different Training Data on the AndroidControl. The evaluation metrics are the action matching score (AMS).}
    \label{tab:androidcontrol_comparison}
    \scalebox{0.65}{
    \begin{tabular}{l| cccc c}
        \toprule
        Training Data & IDD & category\_unseen & app\_unseen & task\_unseen & Overall \\
        \midrule
        AC50k  & 60.43 & 54.46 & 50.00 & 72.10 & 59.25 \\
        AC50k + AITW50k & 60.69 & \textbf{55.26} & 45.19 & 68.84 & 57.50 \\
        AC50k + Ody50k & \textbf{61.48}  & 54.61 & \textbf{50.96} & \textbf{72.46} & \textbf{59.88} \\
        \bottomrule
    \end{tabular}
}
\end{table}

\end{document}